\pdfoutput=1

\documentclass[11pt]{article}

\usepackage{ACL2023}
\usepackage{times}
\usepackage{latexsym}

\usepackage[T1]{fontenc}

\usepackage[utf8]{inputenc}

\usepackage{microtype}

\usepackage{inconsolata}

\usepackage{tikz}
\usepackage{sansmath}
\usepackage{textcomp}
\usepackage{pgfplots}
\pgfplotsset{
layers/my layer set/.define layer set={
            background,
            main,
            foreground
        }{
},
set layers=my layer set,
    }

\usepackage{color,colortbl}
\usepackage{bm}
\usepackage{algorithm}
\usepackage{algpseudocode}

\usepackage{amsmath,amsfonts,bm}

\def\eqref#1{equation~\ref{#1}}

\def\1{\bm{1}}

\def\ve{{\bm{e}}}

\def\vq{{\bm{q}}}

\def\vw{{\bm{w}}}
\def\vx{{\bm{x}}}
\def\vy{{\bm{y}}}
\def\vz{{\bm{z}}}

\DeclareMathAlphabet{\mathsfit}{\encodingdefault}{\sfdefault}{m}{sl}
\SetMathAlphabet{\mathsfit}{bold}{\encodingdefault}{\sfdefault}{bx}{n}

\newcommand{\softmax}{\mathrm{softmax}}

 \definecolor{coolgrey}{rgb}{0.43, 0.43, 0.43}
\definecolor{}{rgb}{0.0, 0.44, 1.0}
\usepackage{booktabs,amsfonts,dcolumn}
\newcommand\tf[1]{\textbf{#1}}

\newcommand{\ba}{$_\texttt{base}$}
\newcommand{\cls}{{\tt[CLS]}}
\newcommand{\pad}{{\tt[PAD]}}
\newcommand{\avg}{{(avg.)}}
\newcommand{\viewA}{1}
\newcommand{\viewB}{2}
\usepackage{preview}
\usepackage{xspace}
\newcommand{\ours}{\emph{mi}\texttt{{CSE}}\xspace}
\usepackage{microtype}
\usepackage{amsmath}
\usepackage{arydshln}
\usepackage{multirow}
\usepackage{graphicx}
\usepackage{subcaption}
\usepackage[colorinlistoftodos]{todonotes}
\usepackage{comment}
\usepackage{bold-extra}
\usepackage{mathtools,tikz,caption}
\DeclareRobustCommand\sampleline[1]{\tikz\draw[#1] (0,0) (0,\the\dimexpr\fontdimen22\textfont2\relax)
  -- (1em,\the\dimexpr\fontdimen22\textfont2\relax);}

\newcommand{\blueline}{\raisebox{2pt}{\tikz{\draw[-,black!1!blue,solid,line width = 1.5pt](0,0) -- (5mm,0);}}}

\newcommand{\redline}{\raisebox{2pt}{\tikz{\draw[-,black!1!red,solid,line width = 1.5pt](0,0) -- (5mm,0);}}}

\newcommand{\violetline}{\raisebox{2pt}{\tikz{\draw[-,violet,solid,line width = 1.5pt](0,0) -- (5mm,0);}}}

\newcommand{\greenline}{\raisebox{2pt}{\tikz{\draw[-,black!1!green,solid,line width = 1.5pt](0,0) -- (5mm,0);}}}

\newcommand{\orangeline}{\raisebox{2pt}{\tikz{\draw[-,black!1!orange,solid,line width = 1.5pt](0,0) -- (5mm,0);}}}

\newcommand*\fullcirc{\tikz\fill (0,0) circle (0.65ex);} 
\definecolor{Gray}{gray}{0.9}

\title{\emph{mi}\texttt{{CSE}}: Mutual Information Contrastive \\Learning for Low-shot Sentence Embeddings}

\author{Tassilo Klein \\
  SAP AI Research \\
  \texttt{tassilo.klein@sap.com} \\\And
  Moin Nabi \\
    SAP AI Research \\
  \texttt{m.nabi@sap.com} \\}

\begin{document}
\maketitle
\begin{abstract}
This paper presents \emph{mi}\texttt{CSE}, a mutual information-based contrastive learning framework that significantly advances the state-of-the-art in few-shot sentence embedding.
The proposed approach imposes alignment between the attention pattern of different views during contrastive learning. Learning sentence embeddings with \emph{mi}\texttt{CSE} entails enforcing the structural consistency across augmented views for every sentence, making contrastive self-supervised learning more sample efficient. As a result, the proposed approach shows strong performance in the few-shot learning domain. While it achieves superior results compared to state-of-the-art methods on multiple benchmarks in few-shot learning, it is comparable in the full-shot scenario. This study opens up avenues for efficient self-supervised learning methods that are more robust than current contrastive methods for sentence embedding.\footnote{Source code and pre-trained models are available at: \\ \url{https://github.com/SAP-samples/acl2023-micse/}}
\end{abstract}

\section{Introduction}
 
Measuring sentence similarity has been challenging due to the ambiguity and variability of linguistic expressions. The community's strong interest in the topic can be attributed to its applicability in numerous language processing applications, such as sentiment analysis, information retrieval, and semantic search~\cite{PILEHVAR201595,iyyer-etal-2015-deep}. 
Language models perform well on these tasks but typically require fine-tuning on the downstream task and corpora ~\cite{reimers-gurevych-2019-sentence,devlin2018bert,pfeiffer2020AdapterHub,mosbach2021on}.
In terms of sentence embeddings, contrastive learning schemes have already been adopted successfully~\cite{vanDenOord2018,liu-etal-2021-fast,gao2021simcse,Carlsson2021ICLR}. The idea of contrastive learning is that positive and negative pairs are generated given a batch of samples. Whereas the positive pairs are obtained via augmentation, negative pairs are often created by random collation of sentences. Following the construction of pairs,
contrastive learning forces the network to learn feature representations by pushing apart different samples (negative pairs) or pulling together similar ones (positive pairs). 
While some methods seek to optimize for selecting ``hard'' negative for negative pair generation~\cite{zhou2022debiased}, others investigated better augmentation techniques for positive pair creation. In this regard, many methods have been proposed to create augmentations to boost representation learning.  
Standard approaches for the augmentation aim at input \emph{data level} (a.k.a \emph{discrete} augmentation), which comprises word level operations such as swapping, insertion, deletion, and substitution~\cite{XieICLR2017,Coulombe2018TextDA,wei-zou-2019-eda}. In contrast to that, \emph{continuous} augmentation operates at the \emph{representation level}, comprising approaches like interpolation or ``mixup'' on the embedding space~\cite{chen-etal-2020-mixtext, cheng-etal-2020-advaug, Guo2019AugmentingDW}. Most recently, augmentation was also proposed in a more continuous fashion operating in a \emph{parameter level} via simple techniques such as drop-out~\cite{gao2021simcse,liu-etal-2021-fast,klein2022scd} or random span masking~\cite{liu-etal-2021-fast}. The intuition is that ``drop-out'' acts as minimal data augmentation, providing an expressive \emph{semantic variation}. 
However, it will likely affect \emph{structural alignment} across views. Since positive pairs are constructed from identical sentences, we hypothesize that the structural dependency over the views should be preserved by utilizing drop-out noise. 
Building on this idea, we maximize the \emph{structural dependence} by enforcing distributional similarity over the attention values across the augmentation views. 
To this end, we employ maximization of the mutual information (MI) on the attention tensors of the positive pairs.
However, since attention tensors can be very high-dimensional, computing MI can quickly become a significant burden if not intractable. 
This paper proposes a simple solution to alleviate the computational burden of MI computation, which can be deployed efficiently. Similar to ~\cite{NEURIPS2020_bcff3f63}, we adopt the Log-Normal distribution to model attention. Empirical evidence confirms this model as a good fit while facilitating the optimization objective to be defined in closed form. In this case, mutual information can be provably reformulated as a function of correlation, allowing native GPU implementation. As discussed above, the proposed approach builds upon the contrastive learning paradigm known to suffer from model collapse.
This issue becomes even more problematic when enforcing MI on the attention level, as it tightens the positive pairs via regularizing the attention. Therefore the selection of negative pairs becomes more critical in our setup. To this end, we utilize momentum contrastive learning to generate harder negatives~\cite{MoCo_He2020}. A ``tighter'' binding on positive pairs and repulsion on ''harder'' negative pairs empowers the proposed contrastive objective, yielding more powerful representations.

Combining ideas from momentum contrastive learning and attention regularization, we propose \ours, a conceptually simple yet empirically powerful method for sentence embedding, with the goal of integrating semantic and structural information of a sentence in an information-theoretic and Transformer-specific manner. We conjecture the relation between attention maps and a form of syntax to be the main driver behind the success of our approach. We speculate that our proposed method injects structural information into the model as an inductive bias, facilitating representation learning with fewer samples. The adopted structural inductive biases provide a ``syntactic'' prior as an implicit form of supervision during training~\cite{wilcox2020structural}, which promotes few-shot learning capabilities in neural language models. To validate this, we introduced a low-shot setup for training sentence embeddings. In this benchmark, we finetune the language model \emph{only} with a small number of training samples. Note that this is a very challenging setup. The inherent difficulty can be attributed to the need to mitigate the domain shift in the low-shot self-supervised learning scheme. We emphasize the importance of this task, as in many real-world applications, only small datasets are often available. Such cases include NLP for low-resource languages or expert-produced texts (e.g., medical records by doctors), personalized LM for social media analysis (e.g., personalized hate speed recognition on Twitter), etc. Our proposed method significantly improves over the state-of-the-art in the low-shot sentence embedding benchmark. This is the first work that explores how to combine semantic and structural information through attention regularization and empirically demonstrates this benefit for low-shot sentence embeddings.

\noindent\textbf{Previous works:} Recently, VaSCL~\cite{zhang-etal-2022-virtual}, ConSERT~\cite{yan-etal-2021-consert}, PCL~\cite{WuPCL22} and ~\cite{chuang2022diffcse} proposed contrastive representation learning with diverse augmentation strategies on positive pair. However, we proposed a principled approach for enforcing \emph{alignment} in positive pairs at contrastive learning without discretely augmenting the data. Similar to us, ESimCSE~\cite{wu2021esimcse} and MoCoSE~\cite{cao-etal-2022-exploring} proposed to exploit a momentum contrastive learning model with negative sample queue for sentence embedding to boost \emph{uniformity} of the representations. However, unlike us, they do not enforce any further tightening objective on the positive pairs nor consider few-shot learning. Very recently, authors in InforMin-CL~\cite{chen2022informin-cl} and InfoCSE~\cite{WuInfoCSE2022}proposed information minimization-based contrastive learning. Specifically, the authors propose to minimize the information entropy between positive embeddings generated by drop-out augmentation. Our model differs from this paper and the method in \cite{bachman2019learning,yang2021mutual,zhang-etal-2020-unsupervised,sordoni2021decomposed,wu2020mutual}, which focuses on using mutual information for self-supervised learning. A key difference compared to these methods is that they estimate MI directly on the representation space. In contrast, our method computes the MI on attention. Other related work include~\cite{zhang2022unsupervised,zhou-etal-2022-debiased,zhang2022contrastive,liu2022transencoder}.

The contributions of the proposed work are:
\textbf{First}, we propose to inject structural information into language models by adding an attention-level objective.
\textbf{Second}, we introduce Attention Mutual Information (AMI), a sample-efficient self-supervised contrastive learning.
\textbf{Third}, we introduce low-shot learning for sentence embedding. We show that our method performs comparably to the state-of-the-art in the full-shot scenario and significantly better in few-shot learning.

\section{Method}

\begin{figure*}\centering
   \includegraphics[width=1.0\textwidth]{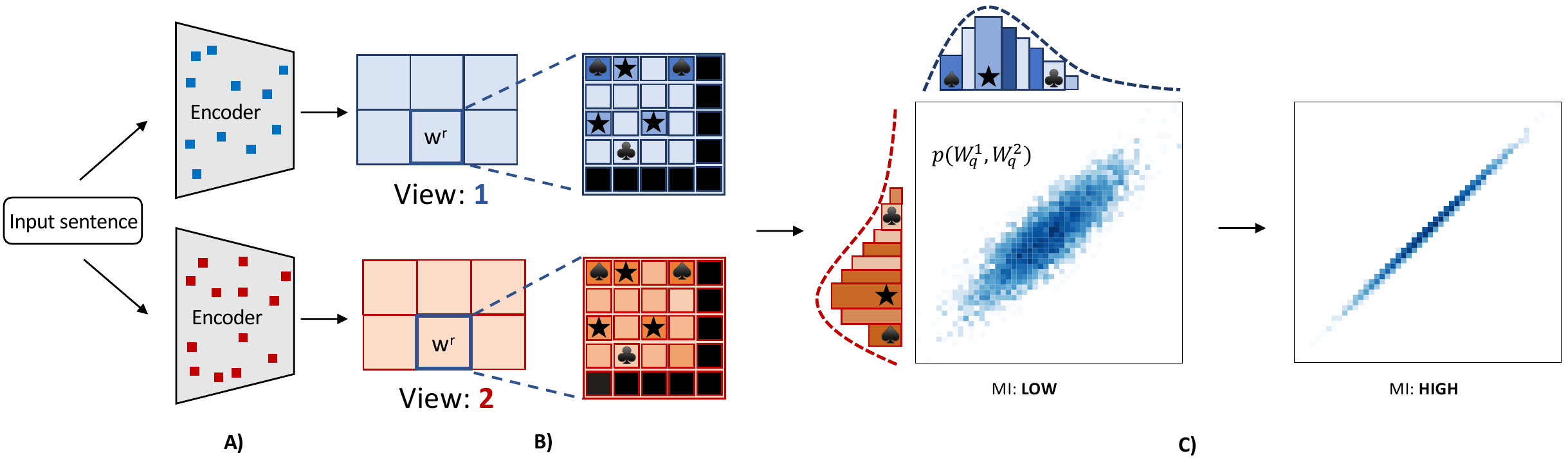}
    \caption{\textbf{Schematic illustration of AMI pipeline:} \textbf{A)} Starting from an input sentence, two views are created by drop-out augmentation (indicated with red and blue). Each view produces a different attention tensor. \textbf{B)} The attention tensor is sliced into tiles, and sampling is then conducted on aligned tiles. High correlation across attention-aligned tiles allows sampling without a significant shift in the attention distribution at a modest accuracy compromise. \textbf{C)} Subsequently, assuming a log-normal distribution of the attention tensor, the joint distribution is computed, and  mutual information is maximized. }
    \label{fig:AMI_pipeline}
\end{figure*}

The proposed approach aims to exploit the structure of the sentences in a contrastive learning scheme. 
Compared to conventional contrastive learning that solely operates at the level of \emph{semantic} similarity in the embedding space, the proposed approach injects \emph{structural} information into the model. This is achieved by regularizing the attention space of the model during training.
We let $\mathcal{D}$ denote a dataset consisting of string sequences (sentences) from corpus $\mathcal{X}$ with $\mathcal{D}=\{x_1,x_2,...,x_{|\mathcal{X}|}\}$, where we assume $x_i$ to be a tokenized sequence of length $n$ with $x_i \in \mathbb{N}^n$. For mapping the input data to the embedding space, we use a bi-encoder $f_\theta$ parametrized by $\theta$. Bi-encoders entail the computation of embeddings for similarity comparison, whereby each sentence in a pair is encoded separately. Hence, the instantiation of a bi-encoder on augmented input data induces multiple views. For the following, we let $v \in \{1,2\}$ denote the index of the view, where each view corresponds to a different augmentation. Consequently, encoding a data batch $\mathcal{D}_b$ yields embedding matrices $E_v \in \mathbb{R}^{|\mathcal{D}_b|\times U}$, where $U$ denotes the dimensionality of the embeddings. Employing a Transformer, encoding the input data yields the embedding matrices and the associated attention tensors $W_v$. 
Then learning representation of the proposed approach entails the optimization of a joint loss:
\begin{eqnarray}
\min_{\theta} \mathcal{L}_{C}(E_\viewA,E_\viewB)+ \mathcal{L}_{D} (W_\viewA,W_\viewB)
\label{eq:loss}
\end{eqnarray}
with $(E_\viewA, W_\viewA),(E_\viewB, W_\viewB)= f_{\theta}(\mathcal{D}_b)$.
Here, $\mathcal{L}_C$ is responsible for the semantic alignment, corresponding to the standard InfoNCE~\cite{vanDenOord2018} loss that seeks to pull positive pairs close together while pushing away negative pairs in the embedding space. 
In contrast, $\mathcal{L}_D$ is responsible for the syntactic alignment, operating on the attention space. However, in comparison to  $\mathcal{L}_D$ is employed only on positive pairs' attention tensors.

\subsection{Embedding-level Momentum-Contrastive Learning (InfoNCE)}
The InfoNCE-loss seeks to pull positive pairs together in the embedding space while pushing negative pairs apart.
Specifically, InfoNCE on embeddings pushes for the similarity of each sample and its corresponding augmented embedding. Negatives pairs are constructed in two ways, reflected by the two terms in the denominator of Eq.~\ref{eq:infonce}. First, in-batch negative pairs are constructed by pairing each sentence with another random sentence (sharing no semantic similarity), pushing for dissimilarity. Second, using embeddings obtained from a momentum encoder known as MoCo~\cite{MoCo_He2020,cao-etal-2022-exploring}. The momentum encoder is a replication of the encoder $f_\theta$, whose parameters are updated more slowly. Specifically, while the parameters of $f_\theta$ encoder are updated via back-propagation, the parameters of the momentum encoder are updated using an exponential moving average from the former. The negative embeddings are produced from samples from previous batches, which are stored in queue $\mathcal{Q}$ and are forward-passed through the momentum encoder. Then the InfoNCE~\cite{vanDenOord2018} loss $(\mathcal{L}_C)$ is defined as:
\begin{equation}
     - \sum_i^{|\mathcal{D}_b|}\log \frac{d(\ve_i, \mathbf{^+{e}}_i)}{\sum\limits_{j:i\neq j}\limits^{|\mathcal{D}_b|}d(\ve_i, \ve_j) + \sum\limits_{j}\limits^{|\mathcal{Q}|}d(\ve_i, \vq_j)} ,
\label{eq:infonce}
\end{equation}
where $\ve_i \in E_\viewA$ and ${^+\ve}_i \in E_\viewB$ denote the embeddings of different augmentations of $x_i$. Furthermore, $d(\vx,\vy)=\exp(sim(\vx,\vy)/\tau)$ with $sim(.)$ the cosine similarity metric, $q_j$ denoting representations obtained from momentum encoder, and $\tau \in \mathbb{R}$ is a temperature scalar.

\subsection{Attention-level Mutual Information (AMI)}
\textbf{Preliminaries and notations:} 
We first briefly review the attention mechanism and explain the notation used in the rest of this section.  
A Transformer stack consists of a stack of $L$ layers, with input data cascading up the layer stack. Each layer comprises a self-attention module and a feed-forward network in its simplest form. Passing sentences through the encoder stack entails simultaneous computation of attention weights. These attention weights indicate the relative importance of every token. To this end, key-value pairs are computed for each token of the input sequence within each self-attention module. This entails the computation of three different matrices: key matrix $K$, value matrix $V$, and query matrix $Q$. 
The values of the attention weights $W$ are obtained according to $W = \softmax(f(Q, K)) \in \mathbb{R}^{n \times n}$, where $f(.)$ is a scaled dot-product. Output features are then generated as obtained according to $WV$. 
To attend to different sub-spaces~\cite{NIPS2017_3f5ee243} simultaneously, the attention mechanism is replicated $H$ times, referred to as multi-head attention.
During training the encoder, the self-attention tensors $W$ values are subject to a random deterministic process, with randomness arising due to drop-out. Hence, the proposed approach seeks to optimize structural alignment by maximizing mutual information between the attention tensors $W_v= [\vw_1,...,\vw_{|\mathcal{D}_b|}]$ of the augmentation views. 
We propose a four-step pipeline to regularize the joint attention space. For a schematic illustration of the AMI pipeline, see Fig.~\ref{fig:AMI_pipeline}.
\\

\noindent\textbf{1) Attention Tensor Slicing: } Given that augmentation has different effects on the attention distribution depending on the depth (layer) and the position (head) in the Transformer stack, we propose to slice the attention tensor. Chunking the attention has multiple advantages. On the one hand, this allows for preserving the locality of distribution change. This is important as it can be empirically observed that distribution divergence between views decreases with increasing depth in the encoding stack. On the other hand, restricting the space permits using a simple distributional model such as bivariate distribution compared to a mixture distribution for the whole stack. 

For the sake of economy in notation and avoid notational clutter, we will restrict the attention tensor of a single encoded sample in the following.
\begin{figure}[hb!]\centering
    \includegraphics[width=0.49\textwidth]{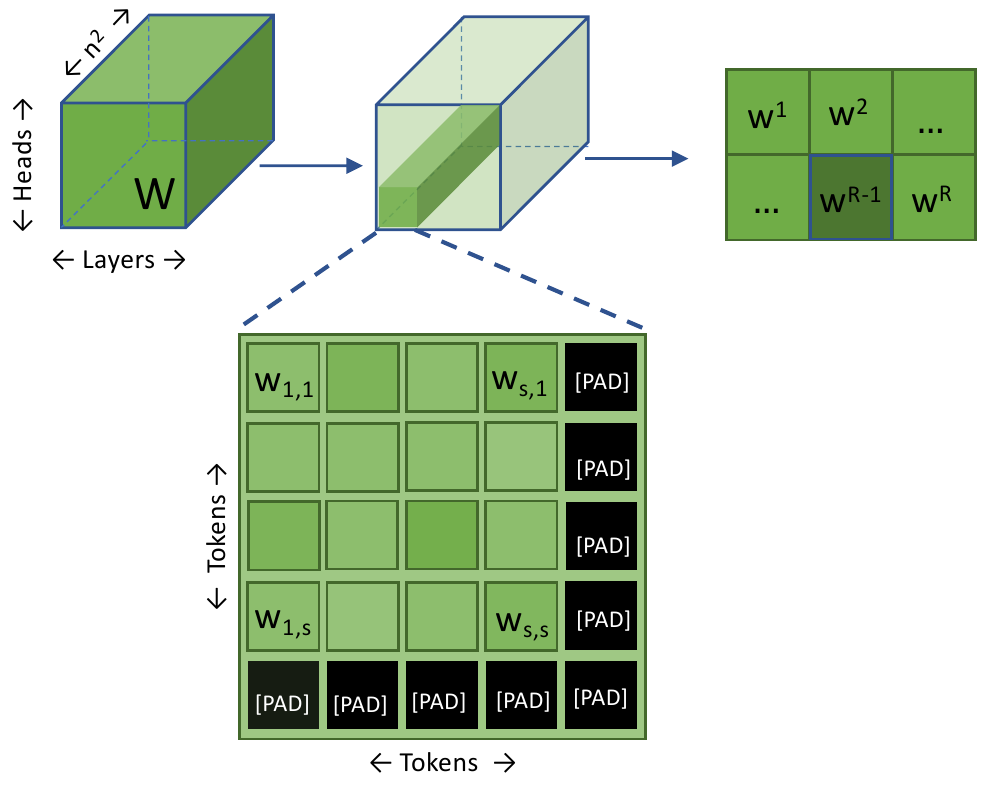} 
    \caption{\textbf{Attention tensor slicing:} Instantiating a transformer stack on an input yields an attention tensor $W$ comprising token attention weights across layers and heads. Slicing the attention entails tiling the tensor. Batch processing of sequences of different lengths is accommodated by padding (\pad).}\label{fig:slicing}
\end{figure}
To this end, a slicing function $\pi: \mathbb{R}^{L \times H \times n \times n}\to\mathbb{R}^{R \times n \times n}$ cuts the attention tensor for each input sample into $R$ (indexed) elements: $\pi(\vw_i) = [\vw_i^1,...,\vw_i^R]\in \mathbb{R}^{n \times n}$ with $\vw_i^{r}  = (w_{j,k})_{1\leq j,k\leq n}$ and $r\in R$. For a schematic illustration of how the attention tensor is sliced into tiles, see Fig.~\ref{fig:slicing}.
\begin{figure}[t!]
    \centering
    \includegraphics[width=0.45\textwidth]{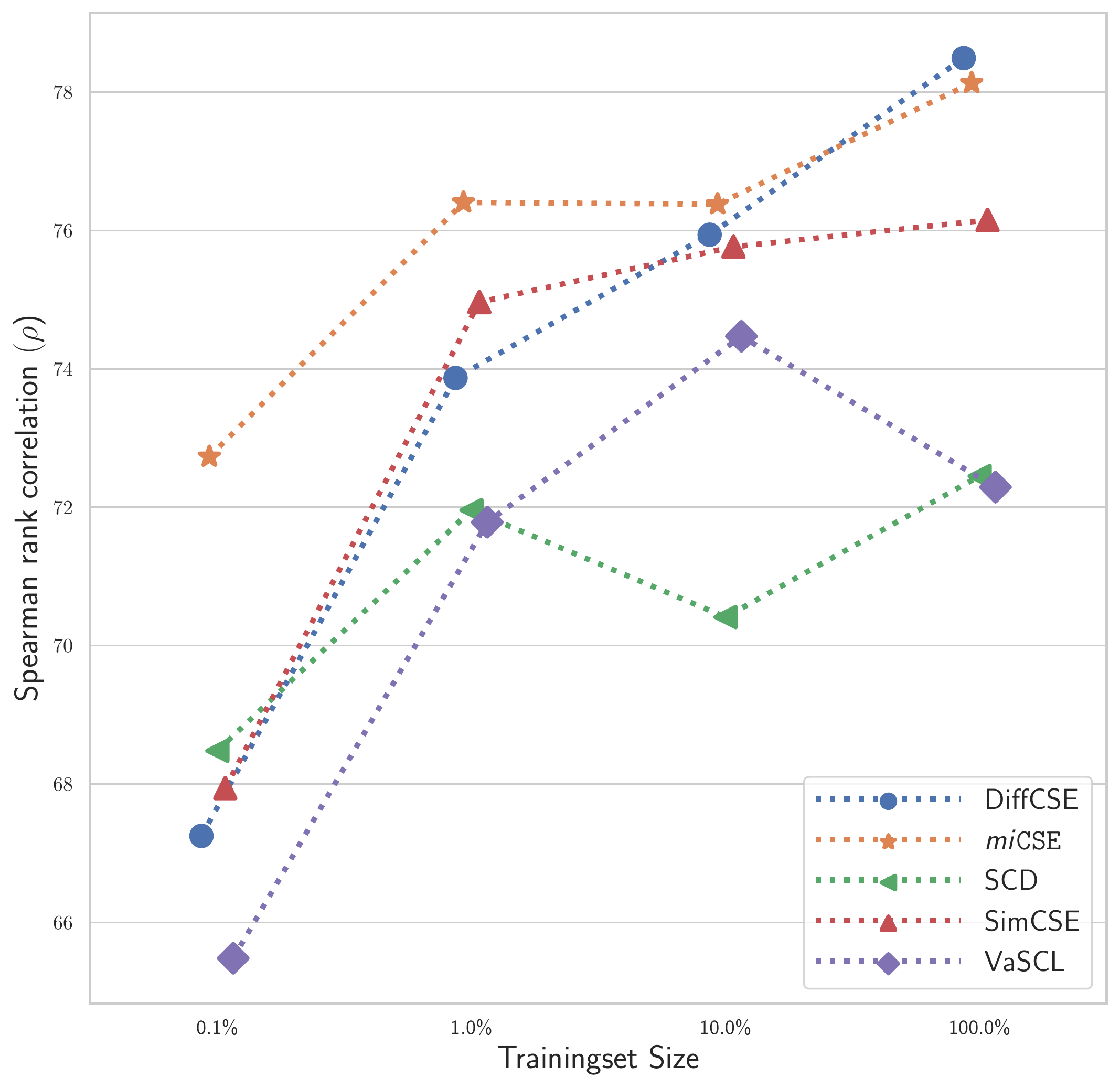}
\caption{Few-shot performance of different algorithms. DiffCSE~\cite{chuang2022diffcse} (\protect\blueline), $\emph{mi}\texttt{{CSE}}$ (\protect\orangeline), SCD~\cite{klein2022scd} (\protect\greenline), SimCSE~\cite{gao2021simcse} (\protect\redline), VaSCL~\cite{zhang-etal-2022-virtual} (\protect\violetline). Performance is shown in Spearman’s correlation average of STS. Training set size: $0.1\%$, $1.0\%$, $10.0\%$, $100.0\%$ of the data. Best viewed in color.} 
    \label{fig:all-methods-fewshot}
\end{figure}
\\

\noindent\textbf{2) Attention Sampling: } 
Different sentences in the batch are typically in token sequences of different lengths. To accommodate the different lengths and facilitate efficient training, sequences are typically padded with \pad-token for length equality. Although this allows for efficient batch encoding on GPU, attentions arising from \pad-tokens have to be discarded when looking at statistical relationships. To accommodate for the different lengths of tokenized sequences, perform a sampling step for attention values within each grid cell $\vw_i^r$. To this end, we leverage multinomial distribution $P_{mult}(p_1,..,p_{n^2})$, where $s$ correspond to the number of non-padding tokens with $1 \leq s \leq n$. Specifically, we sample from the $s^2$ attention values pool, each with a probability of $\frac{1}{s^2}$, with the remaining elements associated with probability $0$. As a result, we obtain a set $J_r=\{j_1,...,j_m\}$ consisting of $m$ indices of the attention tensors for each slice $r\in R$:
\begin{equation}
    J_r \sim P_{mult}(\; \underbrace{1/s^2,...,1/s^2}_\text{$1,..,s^2$},\; \overbrace{0,...,0}^{\mathclap{\text{$(n-s)^2,...,n^2$}}}\;)
\end{equation}
It should be noted that for the same slice $r$ across the views, the same index set is used for sampling: $\tilde{\vw}^r = \bigcup_{j \in J_r}\vw^r[j]$ and ${^+\tilde{\vw}}^{r} = \bigcup_{j \in J_r}{^+\vw}^{r}[j]$.
\\

\noindent\textbf{3) Attention Mutual Information Estimation: } We propose using mutual information to measure the similarity of attention patterns for different views.
Specifically, we follow~\cite{NEURIPS2020_bcff3f63} and adopt the Log-Normal distribution for modeling the attention distribution, which is prudent for several reasons. First, Empirical observation confirms attention asymmetry. Second, by utilizing a non-symmetric distribution, it becomes possible to break down the attention tensor $W$ into $K$ and $Q$, thereby allowing for non-symmetrical attention. Third, adopting the log-normal models facilitates the optimization objective to be defined in closed form and hence easy to optimize, particularly on GPUs.
Mutual information for two normally distributed tuple vectors $(\vz_\viewA, \vz_\viewB)$ can be written as a function of correlation~\cite{MI1957}:\
\begin{equation}
     I(\vz_\viewA, \vz_\viewB) = - \frac{1}{2}\log(1-\rho^2)
     \label{eq:mutual_information}
\end{equation}
where $\rho$ corresponds to the correlation coefficient computed from from $\vz_\viewA$ and $\vz_\viewB$. Hence, we compute the mutual information for each slice $r$ and sample $x_i$ as $MI^r_i = I(\log(\tilde{\vw}_i^r), \log({^+\tilde{\vw}}_i^r))$.
The $log(.)$ function accommodates the Log-Normal to Normal random variable transformation. For details on the implementation, see Alg.~\ref{alg:mutual_information}. 
\\

\noindent\textbf{4) Mutual Information Aggregation: } To compute the loss component for attention regularization, we need to aggregate the distributional similarities for the entire tensor. Aggregation is obtained by averaging the individual similarities obtained for each slice $r \in R$ and each sample $x_i$ in the batch. With $\lambda \in \mathbb{R}$ some weighting scalar, the attention alignment loss term is:
\begin{equation}
    \mathcal{L}_D(W_\viewA,W_\viewB) =  - \frac{\lambda}{|R|\cdot |\mathcal{D}_b|}\sum_{i}^{\mathcal{D}_b}\sum_{r}^R MI^r_i
\end{equation}

\begin{table*}[t!]
    \begin{center}
    \centering
    \small
\begin{tabular}{lcccccccc}
    \toprule

        \multicolumn{9}{c}{\it{Semantic Textual Similarity (STS) Benchmark}}\\
         \midrule
         \tf{Model} & \tf{STS12} & \tf{STS13} & \tf{STS14} & \tf{STS15} & \tf{STS16} & \tf{STS-B} & \tf{SICK-R} & \tf{Avg.} \\
   
    \midrule
         
         BERT & 21.54&	32.11&	21.28&	37.89&	44.24&	20.29&	42.42&	31.40\\ BERT$^\lozenge$(first-last avg) & 39.70 & 59.38 & 49.67 & 66.03 & 66.19 & 53.87 & 62.06 & 56.70 \\
         GloVe$^\clubsuit$(avg.)& 55.14 & 70.66 & 59.73 & 68.25 & 63.66 & 58.02 & 53.76 & 61.32 \\
         BERT-flow$^\lozenge$ & 58.40&	67.10&	60.85&	75.16&	71.22&	68.66&	64.47&	66.55 \\ BERT-whitening$^\lozenge$ & 57.83& 66.90 & 60.90 & 75.08& 71.31& 68.24& 63.73& 66.28\\ 
         
         IS~\cite{zhang-etal-2020-unsupervised} & 56.77 & 69.24 & 61.21 & 75.23 & 70.16 & 69.21 & 64.25 & 66.58 \\
         
         SG-OPT~\cite{kim-etal-2021-self}& 66.84 & 80.13 & 71.23 & {81.56} & 77.17 & 77.23 & 68.16 & 74.62 \\
\hdashline

CT~\cite{Carlsson2021ICLR} & 67.43	& 79.18	& 69.05	& 76.92	& 74.62	& 73.24	& 68.38	& 72.69 \\
        
        SCD$^\dag$~\cite{klein2022scd} & 66.94	& 78.03 &	69.89 &	78.73	& 76.23 & 76.30	& {73.18} &	74.19 \\
        Mirror-BERT$^\dag$~\cite{liu-etal-2021-fast} & 69.10 &	81.10 &	73.00 &	81.90 &	75.70 &	78.00 & 	69.10 &	75.40 \\
         SimCSE~\cite{gao2021simcse} & 68.69 &	82.05 &	72.91 &	81.15 &	79.39 &	77.93 &	70.93 &	76.15 \\
        
        MoCoSE$^\dag$\cite{cao2022exploring} & {71.58} & 81.40 & 74.47 & {{83.45}} & 78.99 & 78.68 & {72.44} & 77.27 \\

        InforMin-CL$^\dag$~\cite{chen2022informin-cl} & 70.22 & {{83.48}} & {75.51} & 81.72 & {79.88} & 79.27 & 71.03 & 77.30 \\

        MixCSE$^\dag$~\cite{zhang2022unsupervised} & 71.71 & 83.14 & 75.49 & 83.64 & 79.00 & 78.48 & 72.19 & 77.66\\
        
\hdashline

ConSERT$_{large}^{\dag,*}$~\cite{yan2021consert} & 70.69 & 82.96 & 74.13 & 82.78 & 76.66 & 77.53 & 70.37 & 76.45 \\

VaSCL$^{\dag,*}$~\cite{wang2022improving} & 69.08 & 81.95 & 74.64 & 82.64 & \tf{{80.57}} & {{80.23}} & 71.23 & 77.19 \\

DCLR$^{\dag,*}$~\cite{zhou2022debiased} & 70.81 & 83.73 & 75.11 & 82.56 & 78.44 & 78.31 & 71.59 & 77.22\\

ArcCSE$^{\dag,*}$~\cite{zhang2022contrastive}& 72.08 & 84.27 & 76.25 & 82.32 & 79.54 & 79.92 & 72.39 & 78.11 \\
        
PCL$^{\dag,*}$~\cite{WuPCL22} & {72.74} & {83.36} & {76.05} & {83.07} & 79.26 & {79.72} & {72.75} & 78.14 \\

ESimCSE$^{\dag,*}$~\cite{wu2021esimcse} & \textbf{73.40} & 83.27 & \textbf{77.25} & 82.66 & 78.81 & 80.17 & 72.30 & {78.27} \\
DiffCSE$^{\dag,*}$~\cite{chuang2022diffcse}& {72.28} & \textbf{84.43} & {76.47} & \textbf{83.90} & {80.54} & \textbf{80.59} & 71.29 & \textbf{78.49} \\
    \midrule

\rowcolor{Gray}
         {\ours} & {71.71} &	{83.09} &	{75.46} &	{83.13} &	{80.22} &	{79.70} &	\tf{{73.62}} &	{78.13} \\
       
\bottomrule
    \end{tabular}
\end{center}
\caption{
        Sentence embedding performance on STS tasks is measured as Spearman’s correlation using BERT\ba, except for VaSCL, which uses RoBERTa. Unless states otherwise, \cls-embedding was used. 
        $\clubsuit$: results from \cite{reimers-gurevych-2019-sentence};
$\lozenge$ results from \cite{gao2021simcse}; $\dag$ by the respective authors;
        other results are by ourselves, denotes the proposed approach, \textbf{bold} denotes the best result, and $*$ denotes the use of \emph{discrete augmentation}.
    }
    \label{tab:main_sts}
\end{table*}

\begin{figure*}[htb]
  \centering
  \begin{minipage}{.85\linewidth}
\begin{algorithm}[H]
\caption{Mutual Information estimation}\label{alg:MI_estimator}
\begin{algorithmic}
\State \textbf{Input:} Batch $\mathcal{D}_b$, encoder {$f_\theta$}, multinomial sampler ${p}_{mult}$
\State \textbf{Output:} Average  mutual information $\frac{1}{|R|\cdot |\mathcal{D}_b|}\sum_{i,r}^{\mathcal{D}_b,R} MI^r_i$
                  
    \State {$(E_\viewA, W_\viewA), (E_\viewB, W_\viewB) \gets {f_\theta}(\mathcal{D}_b)$} \Comment{Transformer encoding creating views}
\For{$i \gets 1...|\mathcal{D}_b|$} 
\State {$\vw_i,^{+}\vw_i \gets \Call{extract}{W_1,W_2,i}$}\Comment{ Extract attention tensor for each sample}
        \State {$\{^{(+)}\vw_i^{1},...,{^{(+)}\vw_i^{R}}\} \gets \pi(^{(+)}\vw_i)$}\Comment{Slicing the attention tensors}
        \State {$s \gets \text{number of text tokens in $x_i$}$}
        \For{$r \gets 1...|R|$}  
            \State {$J_r \gets {p}_{mult}(1/s^2,...1/s^2, \mathbf{0})$} \Comment{Sampling indices of valid attentions}
            \State {$MI^r_i \gets \Call{AMI} { \bigcup_{j \in J_r}\vw_i^r[j],\bigcup_{j \in J_r}{^+\vw_i^{r}}[j]}$}
        \EndFor
    \EndFor

\Procedure{AMI}{$\vw_\viewA,\vw_\viewB$}
    \State {$\vz_\viewA, \vz_\viewB \gets \log(\vw_\viewA), \log(\vw_\viewB)$} \Comment{Log-Normal to Normal transform}
    \State {$\rho \gets \cos(\vz_\viewA-\bar{\vz}_\viewA, \vz_\viewB-\bar{\vz}_\viewB)$} \Comment{Compute correlation coefficient on centered attentions}
	\State Return {$-\frac{1}{2}(1-\rho^2)$} \Comment{Mutual information for tensor slice}
\EndProcedure
\end{algorithmic}
\label{alg:mutual_information}
\end{algorithm}
  \end{minipage}
\end{figure*}

\section{Experiments}
In this section, we describe the experimental setting used for the evaluation, present our main results, and discuss different aspects of our method by providing several empirical analyses. 
\subsection{Experimental Setup}
\noindent\textbf{Model and Hyperparameters: } 
Training is started from a pre-trained transformer LM. Specifically, we employ the Hugging Face~\cite{Wolf2019HuggingFacesTS} implementation of BERT\ba.
For each approach evaluated, we follow the same hyperparameters proposed by the authors. 
In the InfoNCE loss, we set $\tau=0.05$.
In order to determine the hyperparameter $\lambda$ a coarse grid search $\{1.0, 0.1,...,1.0e{-5}\}$ was conducted to assess the magnitude. Upon determination, a fine grid search was conducted once with 10 steps.
We set $\lambda=2.5e-3$ for training $100\%$ of the data in a single episode with a batch size of 50 at a learning rate of $3.0e{-5}$ and $250$ warm-up steps. The number of optimization steps is kept constant for training the different dataset sizes. For the training set of size $10^6 (=100\%)$, we train for 1 epoch; for the size of $10^5 (=10\%)$, we train for 10 epochs, etc. The training was conducted using an NVIDIA V100 with a training time of around $1.5h$. The overall GPU budget from experimentation and hyperparameter optimization is estimated to be around 500 GPU/hours.
The momentum encoder is associated with a sample queue of size $|\mathcal{Q}|=384$. 
The momentum encoder parameters are updated with a factor of $0.995$, except for the MLP pooling layer, which is kept identical to the online network.  
Additionally, we increase the drop-out for the momentum encoder network from the default rate $(0.1)$ to $0.3$.

\begin{table*}\centering
\begin{tabular}{lcccc}
   \toprule

        \multicolumn{5}{c}{\it{Semantic Textual Similarity}}\\
         \midrule
\textbf{Model}               & \tf{0.1\%}        & \tf{1\%}          & \tf{10\%}         & \tf{100\%} \\
    \midrule
CT~\cite{Carlsson2021ICLR}             & $68.46 \pm 2.33$ & $66.21 \pm 4.06$ & $72.06 \pm 1.46$ & $72.69$ \\
\rowcolor{Gray}
{AMI}+CT          & $71.12 \pm 1.11$ & $72.20 \pm 0.49$ & $73.20 \pm 0.78$  & $73.55$ \\
Mirror-BERT~\cite{liu-etal-2021-fast}    & $40.13 \pm 5.08$ & $42.17 \pm 1.69$ & $42.47 \pm 3.66$ & $43.32$ \\
\rowcolor{Gray}
{AMI}+Mirror-BERT & $43.99 \pm 1.26$ & $45.26 \pm 2.60$ & $44.72 \pm 1.36$ & $47.48$ \\
Mirror \avg~\cite{liu-etal-2021-fast}    & $71.48 \pm 1.19$ & $71.80 \pm 1.18$ & $70.38 \pm 1.18$ & $69.81$ \\
\rowcolor{Gray}
{AMI}+Mirror-BERT \avg & $71.49 \pm 0.95$ & $72.54 \pm 0.49$ & $70.68 \pm 1.19$ & $71.34$ \\
SimCSE~\cite{gao2021simcse}         & $67.94 \pm 1.16$ & $74.96 \pm 0.65$ & $75.76 \pm 0.24$ & $76.15$ \\
\rowcolor{Gray}
{AMI}+SimCSE      & $\bm{73.85 \pm 0.49}$ & $76.21 \pm 0.28$ & ${76.31 \pm 0.46}$ & {76.88} \\
$\emph{mi}\texttt{{CSE}}$ & $73.68 \pm 0.89$ & $\bm{76.40 \pm 0.48}$ & $\bm{76.38 \pm 0.35}$ & $\bm{78.13}$ \\
\bottomrule
\end{tabular}

\caption{Sentence embedding few-shot learning performance on STS tasks measured as Spearman’s correlation using BERT\ba. Unless states otherwise, \cls-embedding was used, the number corresponds to the average performance, \textbf{bold} denotes best performance, (\textcolor{Gray}{\fullcirc}) denotes the integration of the proposed approach.}
\label{tab:sts-few-shot}
\end{table*}

\noindent\textbf{Data and Evaluation: } 
Following~\cite{gao2021simcse}, we train the model
unsupervised on sentences from Wikipedia. We create random sample sets of different sizes $\{10^6, 10^5, 10^4, 5.0\cdot10^3, 10^3\}$ to train the model in a few-shot learning scenario.
We repeated the training set creation for each size 5 times with different random seeds.

\noindent\textbf{Mutual Information Estimation: } Following the observations in~\cite{voita2019analyzing},
we restrict the computation of the mutual information to the upper part of the layer stack. Specifically, we select the layers between 8 and 12 (= last layer in BERT\ba). To accommodate input sequences of varying lengths and make computation more efficient, we pool together pairs of adjacent heads (without overlap) while preserving the layer separation. From each of the ($4 \times \frac{H}{2}$) chunks of pooled attentions, we random sample $150$ joint-attention pairs for each embedding of the bi-encoder.

\subsection{Experimental Results}
\noindent\textbf{Unsupervised Sentence Embedding:}
\label{sec:unsup_sentence_embedding}
We compare \emph{mi}\texttt{CSE} to previous state-of-the-art sentence embedding methods on STS tasks. For comparisons, we favored comparable architectures (bi-encoder) that facilitate seamless integration of the proposed approach and methods of comparable backbone. We also added methods that employ explicit \emph{discrete augmentation} to provide a full picture of existing techniques for sentence embedding.

For semantic text similarity, we evaluated on 7 STS tasks: ~\cite{agirre-etal-2012-semeval,agirre-etal-2013-sem,agirre-etal-2014-semeval,agirre-etal-2015-semeval,agirre-etal-2016-semeval}, 
STS Benchmark~\cite{cer-etal-2017-semeval}  and
SICK-Relatedness~\cite{marelli-etal-2014-sick}. These datasets come in sentence pairs with correlation labels in the range of 0 and 5, indicating the semantic relatedness of the pairs. Specifically, we employ the SentEval toolkit~\cite{conneau-kiela-2018-senteval} for evaluation.
All our STS experiments are conducted in a \emph{fully unsupervised} setup, not involving any STS training data. The benchmark measures the relatedness of two sentences based on the cosine similarity of their embeddings. The evaluation criterion is Spearman’s rank correlation ($\rho$). For comparability, we follow the evaluation protocol of~\cite{gao2021simcse}, employing Spearman’s rank correlation and aggregation on all the topic subsets.
Results for the sentence similarity experiment are presented in Tab.~\ref{tab:main_sts}.
As can be seen, the proposed approach is slightly lower in terms of average performance than state-of-the-art algorithms such as DiffCSE. However, it should be noted that these aforementioned methods use extensive discrete augmentation techniques, such as word repetition, deletion, and others, while the proposed method in this work does not employ any form of discrete data augmentation. This renders the proposed method more general and less ad-hoc in nature. While it is technically feasible for our method to incorporate discrete augmentation, it was deliberately excluded in this study for the sake of generalization with the intention of further exploration in future research.
A more in-depth analysis shows the best performance on the SICK-R benchmark, where it outperforms the second-best approach SCD by $(+0.44)$ and third-best PCL by $(+0.87)$. We highlight the comparison to the closest method SimCSE, where the proposed approach has an average gain of $(+3.94)$. This improvement is due to the two additional components (i.e., AMI and MoCo) we add to this baseline method. 

 \begin{figure*}[ht!]
    \begin{minipage}{0.5\textwidth}\subfloat[Few-shot Performance\label{fig:detailed-low-shot-simcse}]{
    \includegraphics[width=0.95\textwidth]{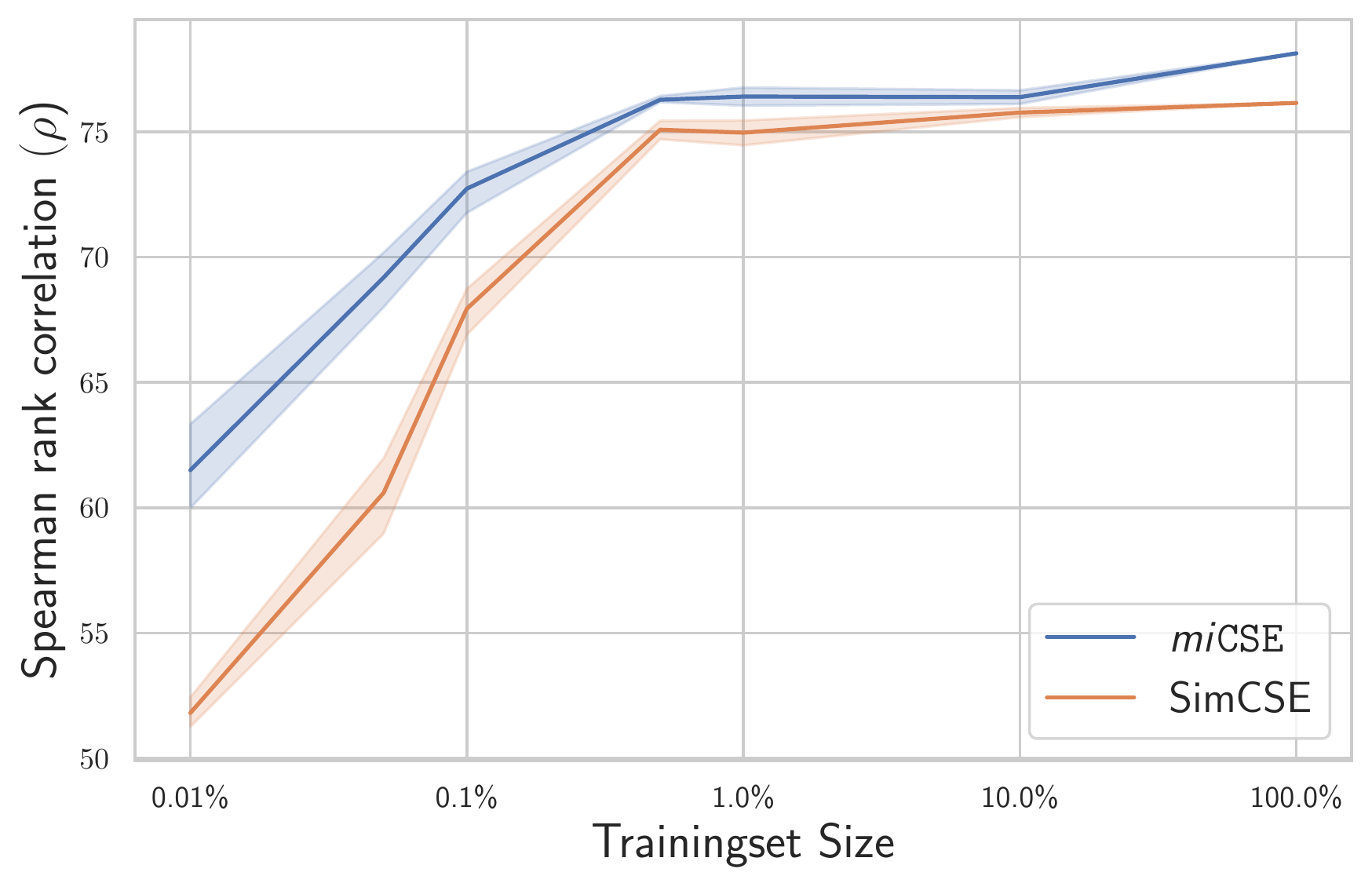}
   }
    \subfloat[Component Analysis\label{fig:delta_component_analysis}]{
     \includegraphics[width=1.0\textwidth]{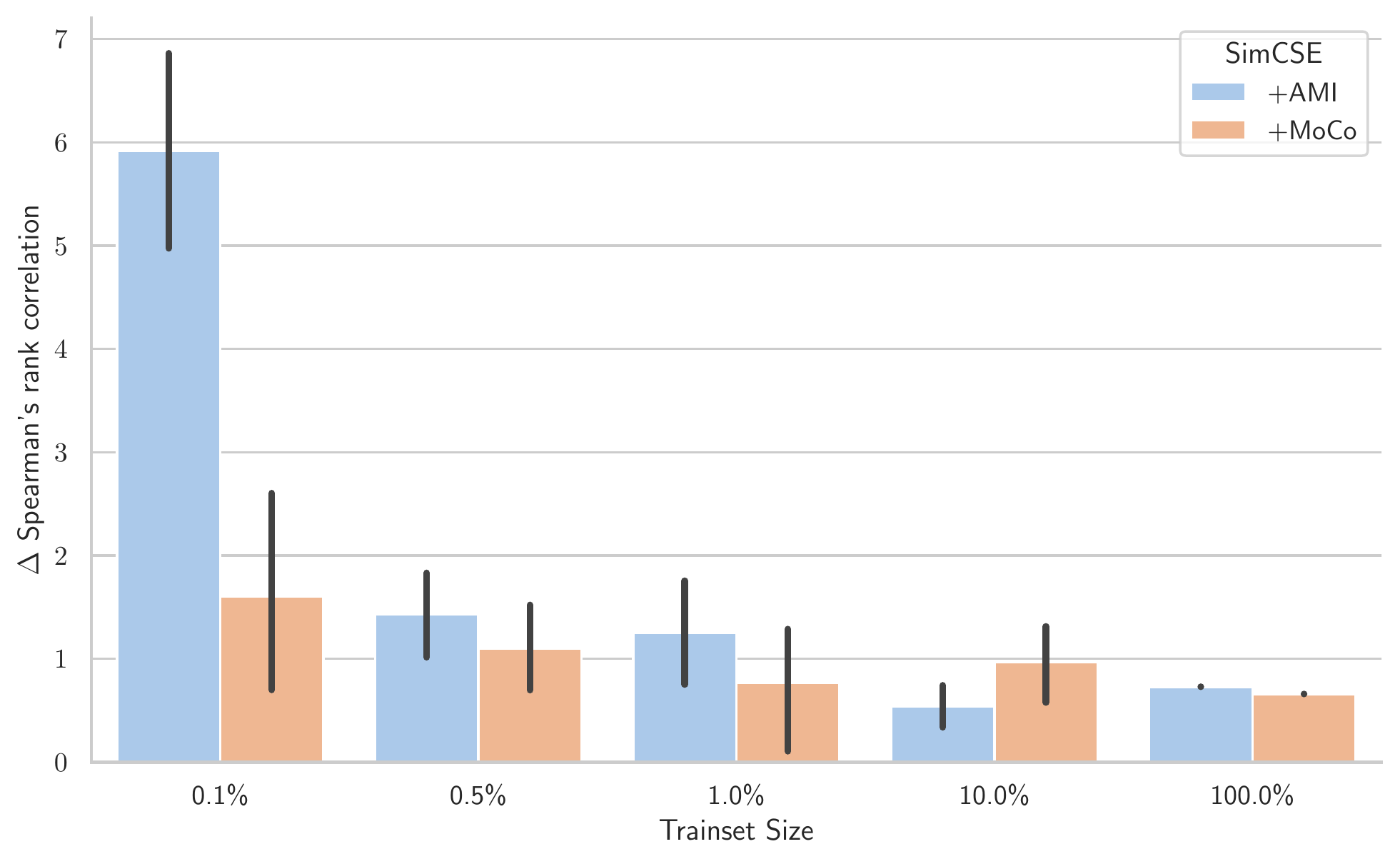}
    }
    \end{minipage}
    \caption{Few-shot performance analysis of models trained with different ratios of dataset size. Performance is shown in Spearman’s correlation average of the STS benchmark. \textbf{Left}: Few-shot performance of SimCSE~\cite{gao2021simcse} (\protect\redline) and the proposed approach $\emph{mi}\texttt{{CSE}}$ (\protect\blueline).  \textbf{Right}: Few-shot ablation study with y-axis showing change ($\Delta$) in Spearman's rank correlation $\rho$, showing the effect of adding components w.r.t. the SimCSE baseline.} 
\end{figure*} 
\noindent\textbf{Low-shot Sentence Embedding:}
In this experiment, the performance of several SOTA sentence embedding approaches is benchmarked elaboratively. 
Similar to Sec.~\ref{sec:unsup_sentence_embedding}, we evaluate 7 STS tasks, STS Benchmark, and SICK-Relatedness with  Spearman’s $\rho$ rank correlation as the evaluation metric. However, in contrast to the previous section, models are trained on different subsets of the data, namely $\{100\%,10\%,1\%,0.1\%\}$ of the Wikipedia dataset used in~\cite{gao2021simcse}.
Results for the low-shot sentence similarity experiment can be presented in Fig.~\ref{fig:all-methods-fewshot}. 
As can be seen, the proposed approach gains by increasing the training set size and consistently outperforms all the baselines in all training subsets. Interestingly, our proposed method reaches the performance of SimCSE trained on the entire dataset with only 0.5\% of the data. We believe it shows the impact of exploiting structural information for data augmentation during training. It should be noted that the performance gain is most significant when conducted on a single token rather than token averaging. We attribute this to token averaging, which to a certain degree, is equivalent to attention regularization.
On the \emph{extremely} low data regime, the proposed approach shows very strong performance up (+11) compared to SimCSE - see Fig.~\ref{fig:detailed-low-shot-simcse}. It suggests resilience of our method to very small batch training.

\subsection{Experimental Analysis of components}

Given that AMI is a regularizer on Transformer attention, we evaluate the applicability in conjunction with other contrastive learning methods. We evaluate the following approaches CT~\cite{Carlsson2021ICLR}, Mirror-BERT~\cite{liu-etal-2021-fast}, and SimCSE~\cite{gao2021simcse}. Evaluation is conducted on 7 STS tasks, STS Benchmark, and SICK-Relatedness with Spearman’s $\rho$ rank correlation as a metric. Results for the low-shot sentence similarity experiment are presented in Tab.~\ref{tab:sts-few-shot}. As can be seen, our proposed AMI can boost the performances of all approaches in all settings.
\begin{table*}[ht!]\centering
\begin{tabular}{lcccc}
   \toprule

        \multicolumn{5}{c}{\it{Semantic Textual Similarity}}\\
         \midrule
\textbf{Model}               & \tf{0.1\%}        & \tf{1\%}          & \tf{10\%}         & \tf{100\%} \\
    \midrule
SimCSE~\cite{gao2021simcse}         & $67.94 \pm 1.16$ & $74.96 \pm 0.65$ & $75.76 \pm 0.24$ & $76.15$ \\
{AMI}+SimCSE      & $\bm{73.85 \pm 0.49}$ & $76.21 \pm 0.28$ & ${76.31 \pm 0.46}$ & {76.88} \\
{MoCo}+SimCSE      & ${69.54 \pm 1.61}$ & ${75.73 \pm 0.91}$ & $\bm{76.73 \pm 0.29}$ & ${76.81}$ \\
$\emph{mi}\texttt{{CSE}}$ & $73.68 \pm 0.89$ & $\bm{76.40 \pm 0.48}$ & $76.38 \pm 0.35$ & $\bm{78.13}$ \\
\bottomrule
\end{tabular}

\caption{Few-shot ablation study using \cls-embedding on STS tasks measured as Spearman’s correlation using BERT\ba.  Performance corresponds to the average across all STS benchmarks, \textbf{bold} denotes best performance. }
\label{tab:sts-few-shot-ablation}
\end{table*}
Additionally, it shows the most significant boost in performance in combination with SimCSE. In addition, we observe that the impact of AMI grows with declining training set size. Combined with SimCSE, AMI leads to a performance gain of up to $(+5.91)$ at $0.1\%$ of the data. We also observe that adding AMI to all the approaches significantly reduces the variance for all methods. This can probably be attributed to the regularization effect of the proposed AMI component. In addition, we conducted an ablation study to assess the effect of AMI and MoCo w.r.t. the baseline SimCSE - see Tab. ~\ref{tab:sts-few-shot-ablation}. As shown in Fig.~\ref{fig:delta_component_analysis}, AMI and MoCo improve the baseline at different data ratios. Again, AMI provides a particularly strong performance boost in the low-data regime. In contrast, the impact of MoCo diminishes with decreasing training set size. We emphasize that our approach gets the best of both worlds by integrating these two components. This can be directly exploited for different few-shot setups by adjusting the hyper-parameter $\lambda$.

\noindent\textbf{Discussion on the \emph{Structure} and \emph{Attention}:}
The proposed approach aligns the attention patterns for drop-out augmented input pairs. We posit that conducting such a regularization enforces constraints w.r.t. the structure (e.g., syntax) of the sentence embeddings. This is motivated by recent literature findings, which suggest that the Transformer\textquotesingle s attention captures structural information such as syntactic grammatical relationships of the sentences~\cite{ravishankar-etal-2021-attention,clark2019does,raganato2018analysis,voita2019analyzing}. 
Additionally, recent research explicitly targets the extraction of topologies from attention maps for diverse tasks on syntactic and grammatical structure~\cite{kushnareva-etal-2021-artificial,Cherniavskii2022,Perez2022}. Although no ``one-to-one'' mapping connects syntactic structures and attention patterns, the attention tensor, at the bare minimum, encodes a ``holistic notion'' of the syntactic structure of sentences. While this study refrains from making any definitive claim on the matter, a preliminary analysis wrt. role of syntax in our proposed method is conducted (see Appendix).

\noindent\textbf{Discussion on the \emph{discrete} argumentation:} 
Discrete augmentation serves as a suitable strategy for expanding datasets to enhance learning robustness and partially address the issue of data scarcity. Although augmentation contributes to improved robustness, additional measures are required to tackle the information gap challenge in few-shot learning scenarios. Therefore, our current study deliberately excluded discrete augmentation to minimize any interference it may have with our low-shot learning algorithm. The primary rationale behind this decision is that while discrete augmentation is known to alleviate data scarcity by replicating missing information, it often leads to a superficial correlation between test and training data, rather than enhancing the model's few-shot learning capability. Consequently, we excluded augmentation to maintain control over \emph{mi}\texttt{CSE}'s behavior and validate its effectiveness without any negative consequences. The significant superiority of miCSE over augmentation-based approaches (such as DiffCSE) in the low-shot setup is evident from Fig. \ref{fig:all-methods-fewshot}. Nevertheless, the proposed approach inherently facilitates the integration of discrete augmentation, offering the potential to enhance results in both few and full-shot learning scenarios. However, it is crucial to acknowledge that their structural similarities must be respected when applying augmentation strategies to positive pairs. One promising option is to utilize the augmentation strategies proposed by ESimCSE~\cite{wu2021esimcse}, which involve word \emph{duplication} and \emph{deletion} to address length biases. This can be followed by enforcing AMI on the shared attention subspaces of the augmented instances. Although we do not explore this approach in our current paper, it presents an intriguing avenue for future research.

\section{Conclusion}
We proposed a method to inject structural similarity into language models for self-supervised representation learning for sentence embeddings. The proposed approach integrates the inductive bias at the level of Transformer attention by enforcing mutual information on positive pairs obtained by drop-out augmentation.
Leveraging attention regularization makes the proposed approach much more sample efficient. Consequently, it outperforms methods with a significant margin in low-shot learning scenarios while having state-of-the-art performance in full-shot to comparable approaches. 
\section{Limitations}
The proposed AMI component is effective in the low-data regime but cannot be generalized to all cases. Future work will investigate the role of syntax in the structural regularization of attention and the extension of the proposed approach to discrete augmentation.

\bibliography{bibliography}
\bibliographystyle{acl_natbib}

\appendix
\section{Appendix}
In the following sections, we add additional details omitted in the main paper due to space restrictions.
First, we show an analysis of the relationship between syntactic structure and semantics. Next,
we illustrate the cosine similarity distribution according to human judgment (ground truth) in Sec. \ref{sec:cos_sim}.
Next, in Sec.~\ref{sec:2dhist}, we visualize the 2D histogram of joint distributions between views.
In Sec.~\ref{sec:non-contrastive}, we present detailed results of the few-shot performance of \ours~ in contrastive and non-contrastive setup.
Finally, the exact relation between mutual information and correlation is presented in Sec. \ref{sec:bivariate_mi}.

\section{Analysis on Structure vs. Semantic}
\begin{figure*}[ht!]
\centering

\begin{minipage}{.49\textwidth}
\begin{preview}
  \centering
\begin{tikzpicture}

\centering
    \begin{axis}[
        width=7.25cm,
        height=5.0cm,
xmin= -0.2,     xmax= 0.65,   ymin= -0.6,     ymax= 0.65, legend style={font=\tiny\sansmath\sffamily, at={(0.97,0.03)},anchor=south east, fill opacity=0.95},
tick align=inside,
        every axis/.append style={font=\tiny\sansmath\sffamily},
        minor tick style={thick},
        scaled y ticks = false,
        ytick={-0.5,0.60},
yticklabels = {Low, High},
        ytick style={draw=none},
        xtick={-0.20,0.60},
        xticklabels = {Low, High},
        xtick style={draw=none},
axis line style = {very thick,shorten <=-0.5\pgflinewidth},
        axis lines = left,
        axis line style = very thick,
        xlabel=Semantic Similarity,
        x label style={at={(axis description cs:0.5,0.075)},
                       anchor=north,
                       font=\boldmath\sansmath\sffamily\tiny},
        ylabel=Syntactic Similarity,
        y label style={rotate=270,at={(axis description cs:0.20,0.5)},
                       anchor=south,
                       rotate=90,
                       font=\boldmath\sansmath\sffamily\tiny},
        ]
\addplot[
scatter,
only marks,
point meta=explicit symbolic,
scatter/classes={
 d={mark=x,red!50!black}, c={mark=o,red!90!black}, b={mark=o,blue!50!black,fill opacity=0.0,draw opacity=0}, a={mark=o,blue!90!black,fill opacity=0.0,draw opacity=0}
} ]
table[col sep=comma, meta=label] {data/data_negatives_small.csv};

\addlegendentry{SimCSE (Dissimilar Syntax)}
\addlegendentry{SimCSE (Similar Syntax)}

 \addplot[only marks,mark=*,mark options={scale=1, fill=black},text mark as node=true] coordinates { (0.0269,-0.191)  (0.13,0.17)};
\addplot [<->, green, line width=0.5mm,on layer=foreground] coordinates {(0.0269,-0.191)  (0.13,0.17)}; \end{axis}
\end{tikzpicture}
\end{preview}
\end{minipage}\begin{minipage}{.49\textwidth}
\begin{preview}
\begin{tikzpicture}
    \begin{axis}[
        width=7.25cm,
        height=5.0cm,
xmin= -0.2,     xmax= 0.65,   ymin= -0.6,     ymax= 0.65, legend style={font=\tiny\sansmath\sffamily, at={(0.97,0.03)},anchor=south east, fill opacity=0.95},
tick align=inside,
        every axis/.append style={font=\tiny\sansmath\sffamily},
        minor tick style={thick},
        scaled y ticks = false,
        ytick={-0.5,0.60},
yticklabels = {Low, High},
        ytick style={draw=none},
        xtick={-0.15,0.6},
        xticklabels = {Low, High},
        xtick style={draw=none},
axis line style = {very thick,shorten <=-0.5\pgflinewidth},
        axis lines = left,
        axis line style = very thick,
        xlabel=Semantic Similarity,
        x label style={at={(axis description cs:0.5,0.075)},
                       anchor=north,
                       font=\boldmath\sansmath\sffamily\tiny},
        ylabel=Syntactic Similarity,
        y label style={rotate=270,at={(axis description cs:0.20,0.5)},
                       anchor=south,
                       rotate=90,
                       font=\boldmath\sansmath\sffamily\tiny},
        ]
\addplot[
scatter,
only marks,
point meta=explicit symbolic,
scatter/classes={
b={mark=x,blue!50!black}, a={mark=o,blue!90!black}, c={mark=x,red!90!black,fill opacity=0.0,draw opacity=0}, d={mark=x,red!50!black,fill opacity=0.0,draw opacity=0}
}, ]
table[col sep=comma, meta=label] {data/data_negatives_small.csv};

\addlegendentry{\emph{mi}\texttt{CSE} (Dissimilar Syntax)}
  \addlegendentry{\emph{mi}\texttt{CSE} (Similar Syntax)}
   \addplot[only marks,mark=*,mark options={scale=1, fill=black},text mark as node=true] coordinates { (0.175,0.331)  (0.0458,-0.339)};
 \addplot [<->, green, line width=0.5mm,on layer=foreground] coordinates {(0.175,0.331)  (0.0458,-0.339)}; \end{axis}
\end{tikzpicture}
\end{preview}
\end{minipage}

\caption{Comparison of negative contrastive pairs sentence embeddings in terms of semantic and syntax. \textbf{Left}: SimCSE  \textbf{Right}: \emph{mi}\texttt{CSE}. Semantic similarity measured in terms of cosine similarity, syntactic similary measured with mutual information on attentions. Negative pairs sub-divided into pairs with similar/dissimilar dependency trees. (\textcolor{black}{\fullcirc}) denote cluster centroids, (\textcolor{green}{\textbf{$\leftrightarrow$}}) distance between centroids. Ranges are aligned.}
\label{fig:hypothesis_negatives}
\end{figure*}
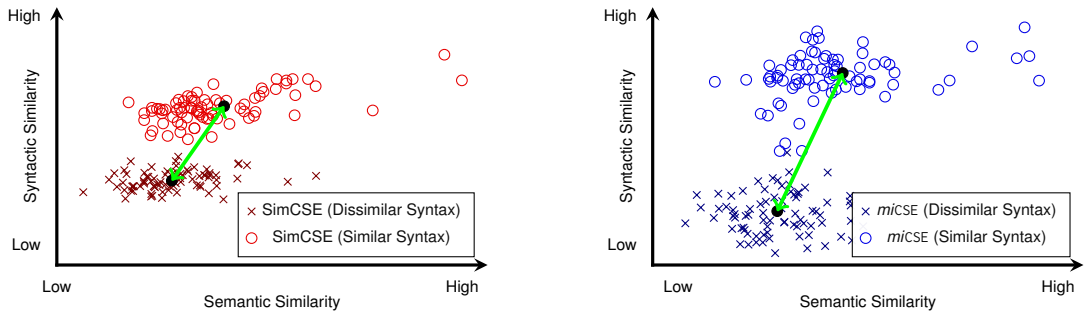

In light of the lack of a rigorous benchmark for analyzing structure(syntax) in sentence embedding, we performed two qualitative analyses visualized in Fig.~\ref{fig:hypothesis_pos_and_neg} and Fig~\ref{fig:hypothesis_negatives}.

Let us consider the following three sentences and their linearized syntax tree to understand better the notions of negatives and (dis-)similar syntax.
\\
\\\textbf{Anchor / Positive:} \\Life is good
\\\textbf{Negative (\texttt{\emph{similar} Syntax}):} \\Good is expensive
\\\textbf{Negative (\texttt{\emph{dissimilar} Syntax}):} \\Live a good life
\\\\
For each sentence, we computed the dependency tree. Subsequently, we linearize the tree structure for comparison, as can be done with tools such as spaCy\footnote{https://spacy.io/}. Positive samples have an identical tree and negative samples have non-identical trees with their part-of-speech tags:
\\
\\\textbf{Anchor / Positive:}  
\\\texttt{nsubj(1,0) -- ROOT(1,1) -- acomp(1,2) -- punct(1,3).}
\\\textbf{Negative (\texttt{\emph{similar} Syntax}):} 
\\\texttt{nsubj(1,0) -- ROOT(1,1) -- acomp(1,2) -- punct(1,3).}
\\\textbf{Negative (\texttt{\emph{dissimilar} Syntax}):} 
\\\texttt{ROOT(0,0) -- det(3,1) -- amod(3,2) -- npadvmod(0,3) -- punct(0,4).}
\\\\
Here \texttt{nsubj} corresponds to "nominal subject," \texttt{acomp} to "adjectival complement," \texttt{det} to "determiner," \texttt{npadv} to "noun phrase as adverbial modifier" and \texttt{punct} to "punctuation."

Our empirical observations are:

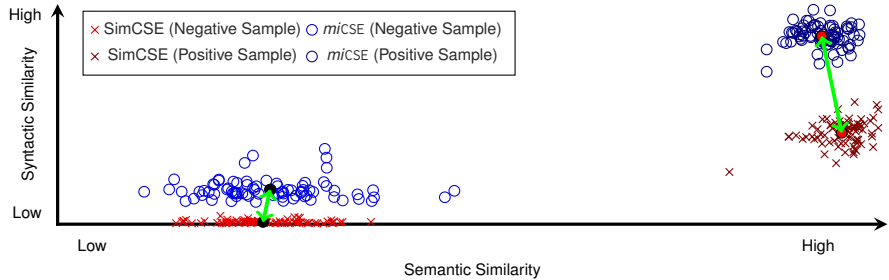
\begin{figure*}[ht!]
\centering
\begin{minipage}{.95\textwidth}
\centering
\begin{preview}
  \centering
\begin{tikzpicture}
    \begin{axis}[
    legend columns = 2,
        width=12.5cm,
        height=4.5cm,
xmin= -0.3,     xmax= 0.9,   ymin= 0.0,     ymax= 1.9, legend style={font=\tiny\sansmath\sffamily,at={(0.03,0.97)},anchor=north west,fill opacity=0.85},
ytick={0.1,1.8},
        yticklabels = {Low, High},
        ytick style={draw=none},
        xtick={-0.25,0.8},
        xticklabels = {Low, High},
        xtick style={draw=none},
        every axis/.append style={font=\tiny\sansmath\sffamily},
        minor tick style={thick},
        scaled y ticks = false,
axis line style = {very thick,shorten <=-0.5\pgflinewidth},
        axis lines = left,
        axis line style = very thick,
        xlabel=Semantic Similarity,
        x label style={at={(axis description cs:0.5,0.05)},
                       anchor=north,
                       font=\boldmath\sansmath\sffamily\tiny},
        ylabel=Syntactic Similarity,
        y label style={rotate=270, at={(axis description cs:0.10,0.5)},
                       anchor=south,
                       rotate=90,
                       font=\boldmath\sansmath\sffamily\tiny},
        ]
\addplot[
scatter,
only marks,
point meta=explicit symbolic, on layer=background,
scatter/classes={
c={mark=x,red!90!black}, a={mark=o,blue!90!black}, d={mark=x,red!50!black}, b={mark=o,blue!50!black}} ]
table[col sep=comma, meta=label] {data/data_positives_small.csv};

\addlegendentry{SimCSE (Negative Sample)}
 \addlegendentry{\emph{mi}\texttt{CSE} (Negative Sample)}
   \addlegendentry{SimCSE (Positive Sample)}
 \addlegendentry{\emph{mi}\texttt{CSE} (Positive Sample)}
 
 \addplot[only marks,mark=*,mark options={scale=1, fill=red},text mark as node=true] coordinates {  (0.805,1.623)  (0.834,0.789)};
 
  \addplot[only marks,mark=*,mark options={scale=1, fill=black},text mark as node=true] coordinates { (0.007,0.298)  (-0.003,0.019) };
 
\addplot [<->, green, line width=0.5mm,on layer=foreground] coordinates {(0.007,0.298)  (-0.003,0.019)}; \addplot [<->, green, line width=0.5mm,on layer=foreground] coordinates {(0.805,1.623)  (0.834,0.789)}; \end{axis}
\end{tikzpicture}
\end{preview}
\end{minipage}

\caption{Sentence embeddings of positive and negative contrastive pairs in terms of semantic and syntax, comparing SimCSE and \emph{mi}\texttt{CSE}. Semantic similarity is measured in terms of cosine similarity, syntactic similarity measured with mutual information on attention-level. (\textcolor{red}{\fullcirc}) and (\textcolor{black}{\fullcirc}) denote centroids of positive and negative centroids, (\textcolor{green}{\textbf{$\leftrightarrow$}}) their distance.}
\label{fig:hypothesis_pos_and_neg}
\end{figure*}  
\textbf{Observation (i)} \emph{There is a higher semantic and syntactic similarity between positive pairs compared to the negative pairs}: Our contrastive learning approach assumes that positive pairs exhibit more syntactic similarity than negative pairs (i.e., syntactic inductive bias). To validate this hypothesis, we plot the semantic similarity against syntactic similarity for both positive and negative pairs. Specifically, we analyzed the embeddings and attention values of the trained model with SimCSE and the proposed approach. Input to the models was randomly sampled sentences from Wikipedia. Interestingly enough, although training the proposed model involves maximization of MI over the attention w.r.t. positive pairs, we also observe the reflection of syntactic information in the negative pairs. As shown in Fig.~\ref{fig:hypothesis_pos_and_neg}, the negative pairs end up in the low left corner, whereas the positive pairs are in the upper right corner. 

\textbf{Observation (ii)}: \emph{Negative pairs with similar syntax show higher attention similarity, compared to pairs with dissimilar syntax}: 
For a more in-depth analysis of this, we further sub-divided the negative pairs into two groups: \emph{a)} negative pairs with similar dependency trees, \emph{b)} negative pairs with dissimilar dependency trees. For simplicity, we adopted a binary similarity scheme - ``similar'' implies an identical dependency tree, whereas ``dissimilar'' corresponds to a non-identical dependency tree. To highlight the inter-group syntax similarity, samples of each group were normalized w.r.t. the centroid of the opposite group. As shown in Fig~\ref{fig:hypothesis_negatives} (by the increased distance between the cluster centers), the proposed approach encodes a notion of syntactic similarity. Note that this margin appeared solely due to enforcing the AMI on attention for the positive pairs, leading to a notion of ``syntax'' on negative pairs.

\begin{figure*}\centering
\begin{minipage}{.99\textwidth}
  \centering
  \includegraphics[width=.99\linewidth]{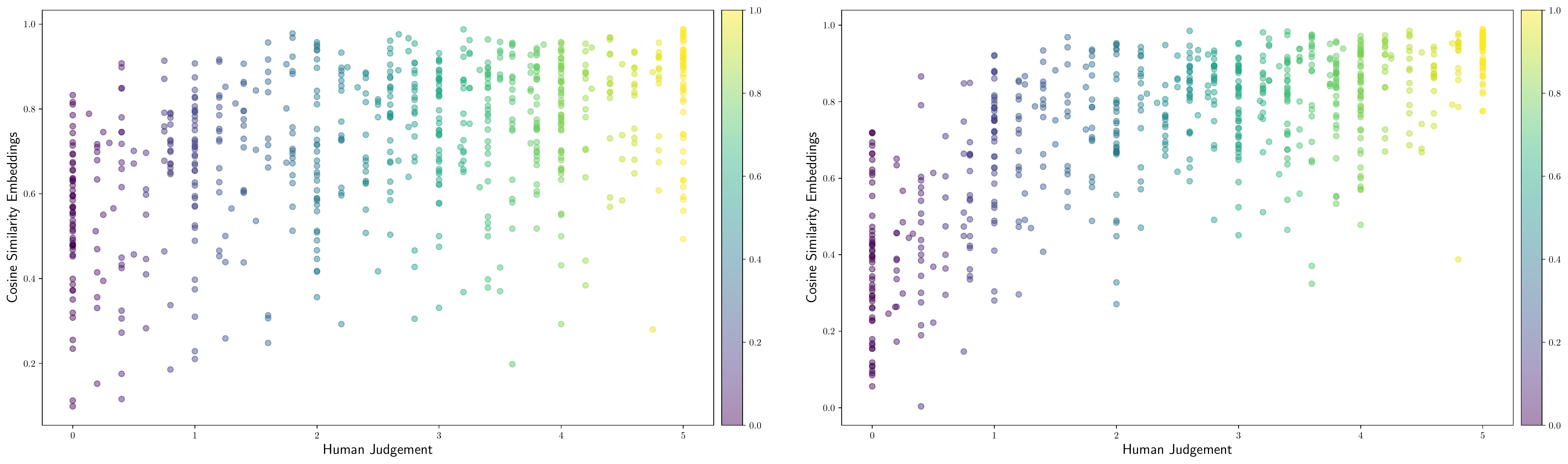}
\end{minipage}\caption{Scatter plots of cosine similarities between sentence pairs in STS. Pairs are shown based on ground-truth human scores (higher means more similar) along the x-axis; the y-axis is the cosine similarity. Color coding corresponds to ground-truth similarity. \textbf{Left:} SimCSE, \textbf{Right:} \ours~ (best viewed in color)}
\label{fig:correlation_analysis}
\end{figure*}

\section{Cosine-similarity Distribution}\label{sec:cos_sim}
To directly show the strengths of our approaches on STS tasks, we illustrate the cosine similarity on embeddings distributions of STS-B pairs in combination with human ratings in Fig. \ref{fig:correlation_analysis}. The STS dataset comes in sentence pairs with correlation labels in the range of 0 and 5, indicating the semantic relatedness
of the pairs. Here, the x-axis is the sample similarity of sentences according to human judgment (ground truth), and the y-axis represents the cosine similarity between pairs using embeddings. Color coding corresponds to ground-truth similarity. Compared to the baseline model (SimCSE), \ours~ better distinguishes sentence pairs with different levels of similarities, as can be seen from the stronger correlation between embedding distance and human rating. This property leads to better performance on STS tasks. In addition, we observe that \ours~ generally shows a more scattered distribution while preserving a lower variance on semantically similar sentence pairs. This observation further validates that \ours~ can potentially achieve a better alignment-uniformity balance.

\section{Visualization of Joint Distribution}\label{sec:2dhist}
To analyze the impact of the proposed approach compared to the baseline SimCSE at the attention level, we visualized the joint distribution of the attention values created by the two views created by the bi-encoder. The joint distribution and mutual information are closely related. More specifically, given two random variables $X$ and $Y$, the associated mutual information can be expressed in terms of the joint distribution as:
\begin{equation}
    I(X,Y) = \sum_{x,y}p(x,y)\log \frac{p(x,y)}{p(x)p(y)},
\end{equation}
where $p(x,y)$ denotes the joint-distribution and $p(x), p(y)$ the marginals. Assuming random variables are normally distributed, the joint distribution of random variables is distinctly shaped depending on the correlation coefficient $\rho$.
See Sec.~\ref{sec:bivariate_mi} details on the relationship between entropy and the correlation coefficient.
In the extreme case of totally unrelated marginals $\rho = 0$, the joint distribution assumes a circular shape having the lowest possible mutual information. On the other end of the spectrum, in the case of perfect correlation, the joint distribution assumes collinearity (45$^{\circ}$ diagonal), with mutual information assuming maximal value. 
We sliced the attention tensor into 12 slices to avoid visual clutter, pooling together every 3 adjacent heads and every 4 adjacent layers. Slicing the tensor at a higher resolution leads to visually very similar results. The axes of the joint distribution (2d histogram) correspond to the marginals' distribution. As \ours~ maximizes the mutual information, one can observe a reduction in the scatter of the joint distribution compared to SimCSE.

\begin{figure*}[t!]
\begin{minipage}{0.33\textwidth}\subfloat{
\includegraphics[width=1.0\textwidth]{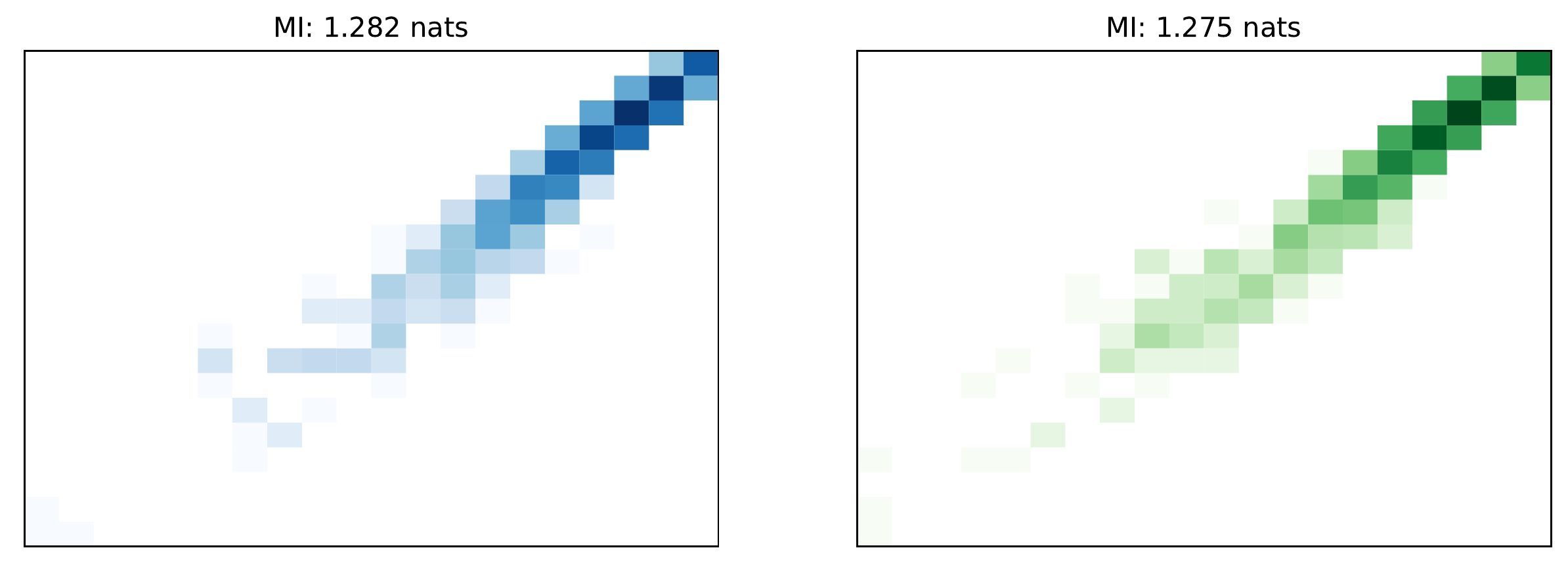}
\label{fig:subfig1}}
\subfloat{
\includegraphics[width=1.0\textwidth]{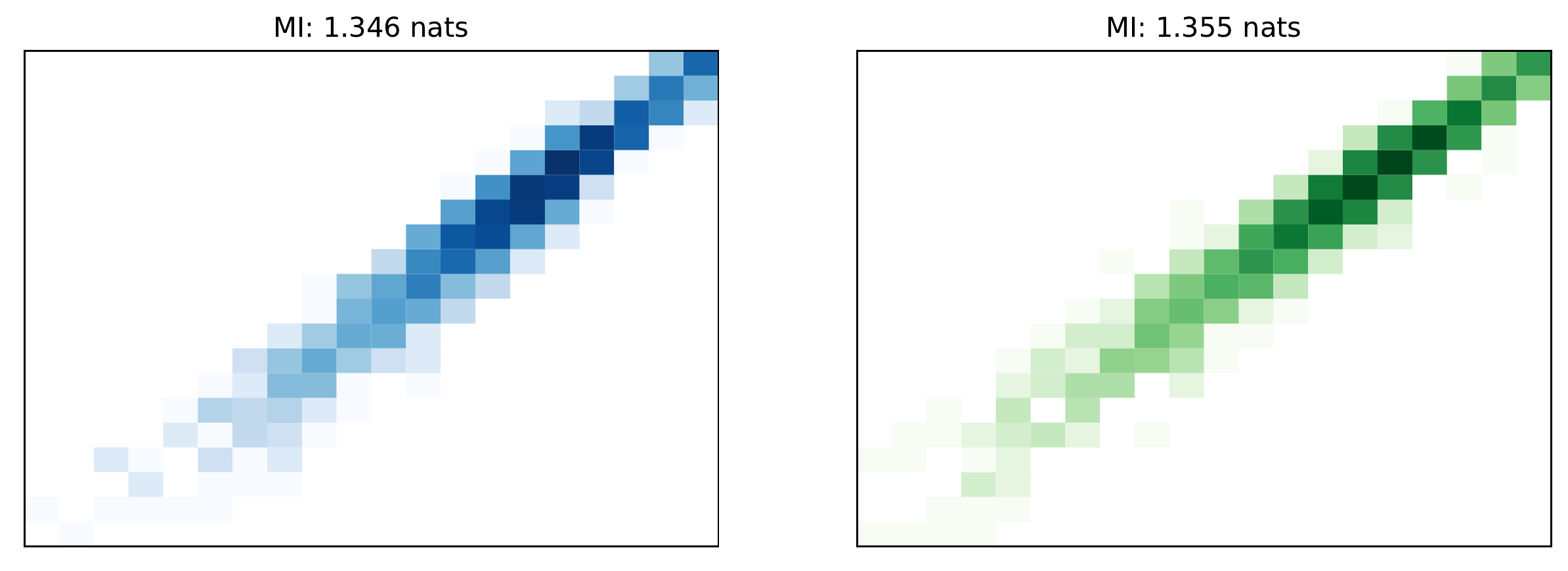}
\label{fig:subfig2}}
\subfloat{
\includegraphics[width=1.0\textwidth]{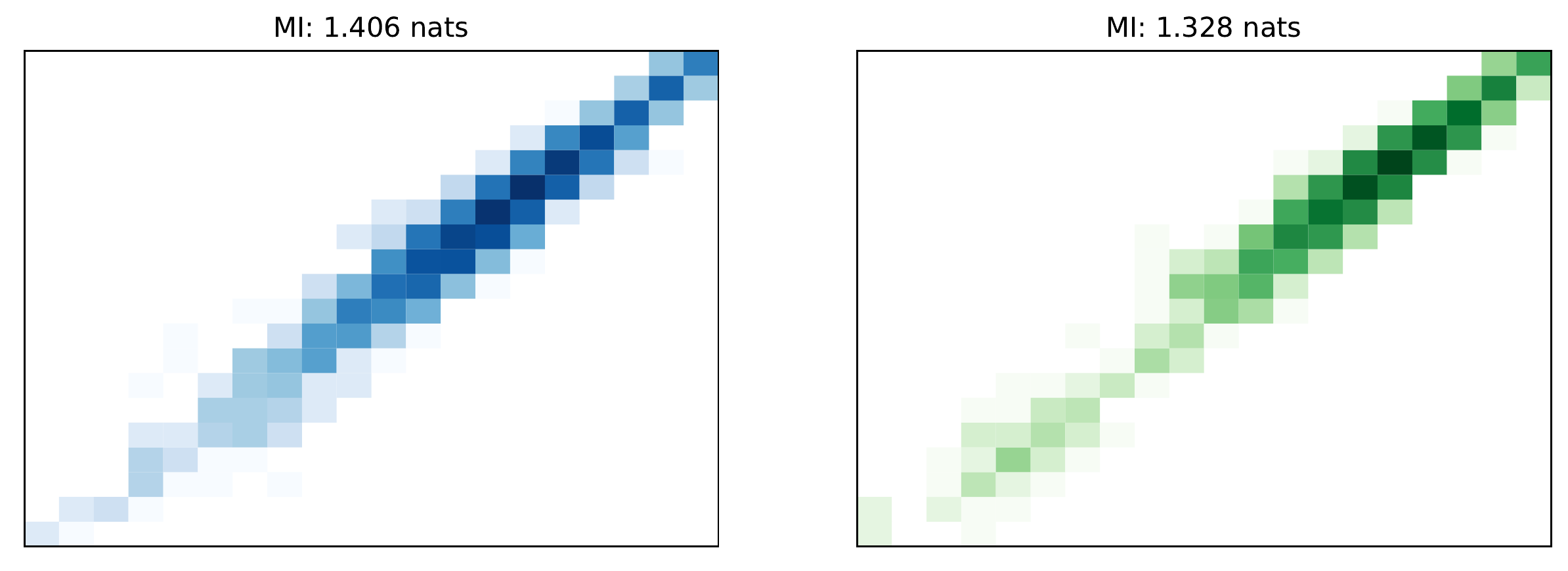}
\label{fig:subfig3}} \\
\subfloat{
\includegraphics[width=1.0\textwidth]{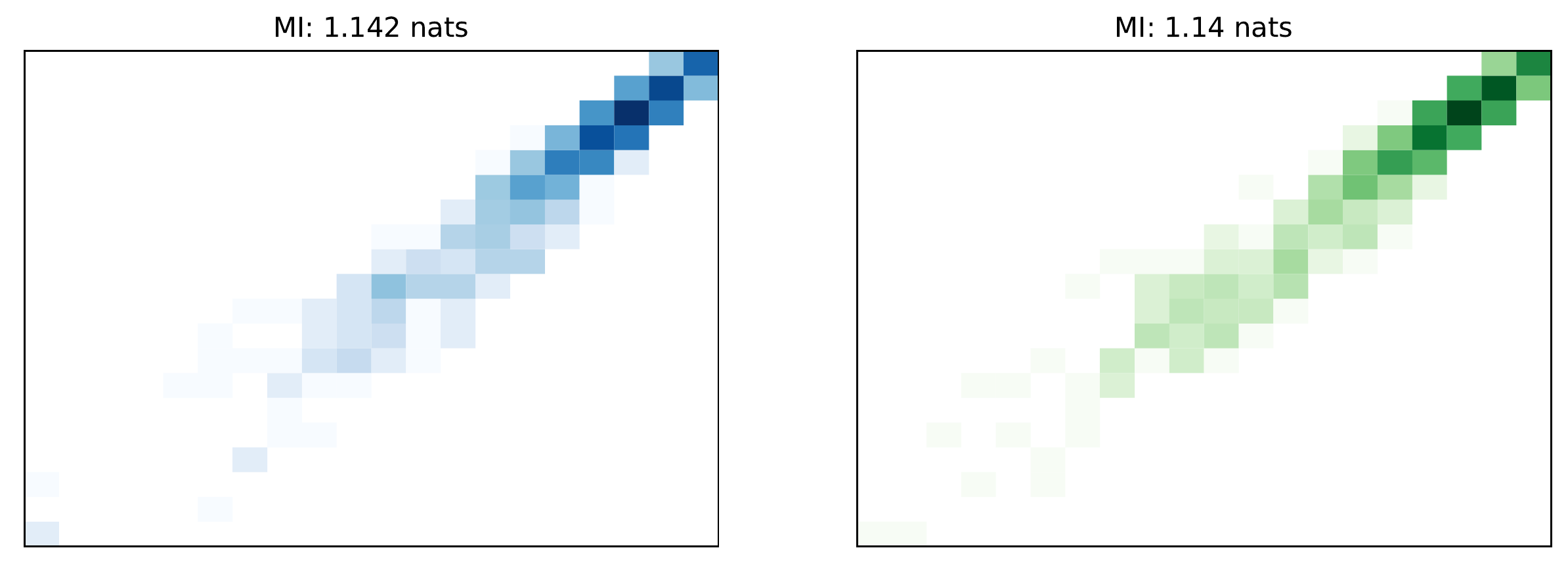}
\label{fig:subfig4}}
\subfloat{
\includegraphics[width=1.0\textwidth]{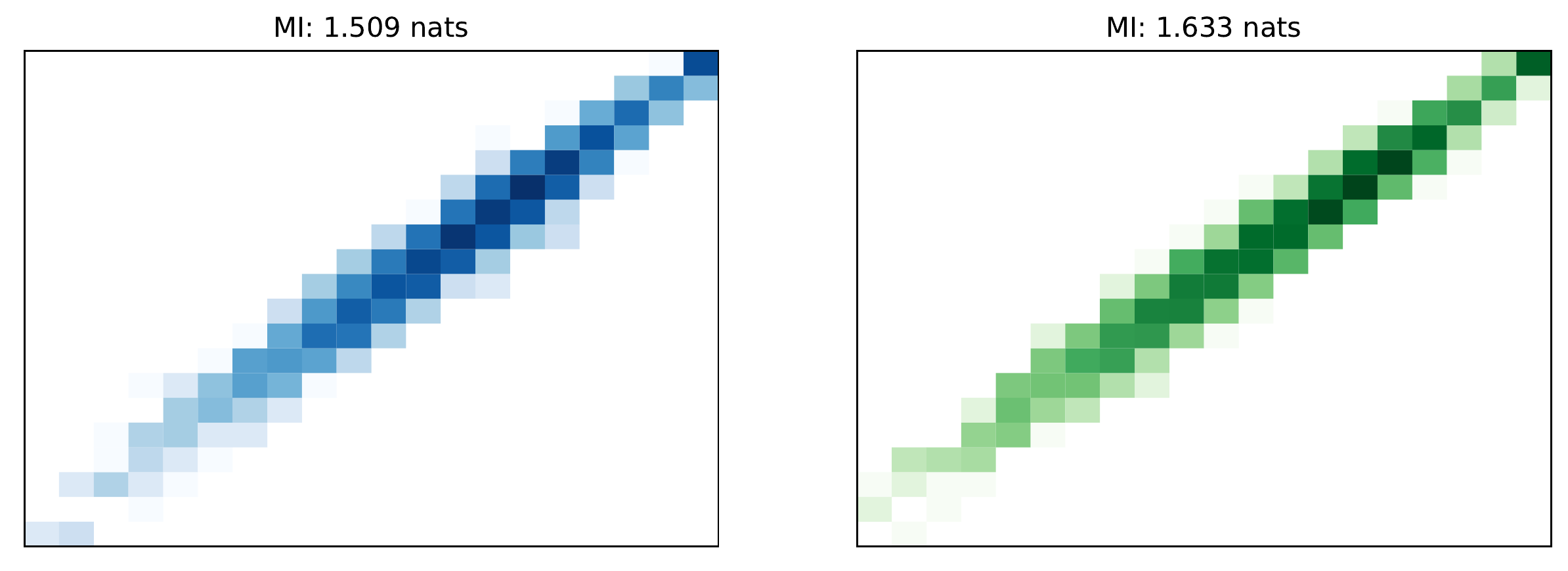}
\label{fig:subfig5}}
\subfloat{
\includegraphics[width=1.0\textwidth]{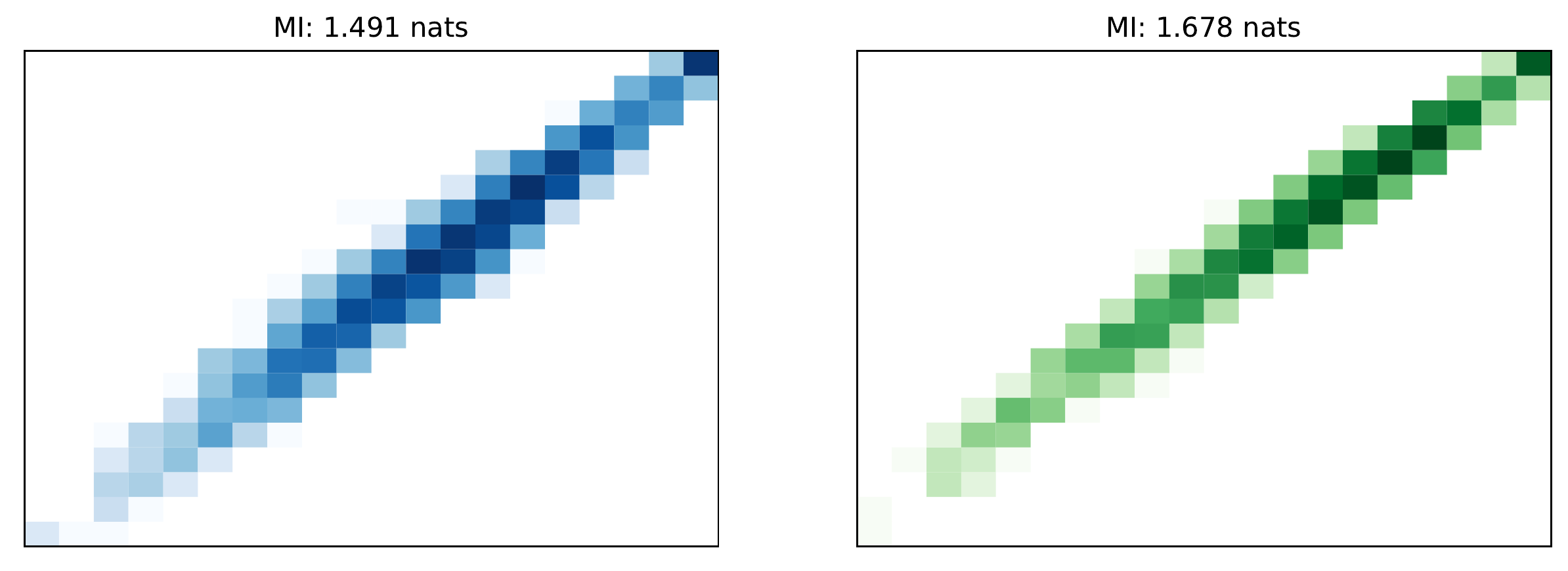}
\label{fig:subfig6}} \\
\subfloat{
\includegraphics[width=1.0\textwidth]{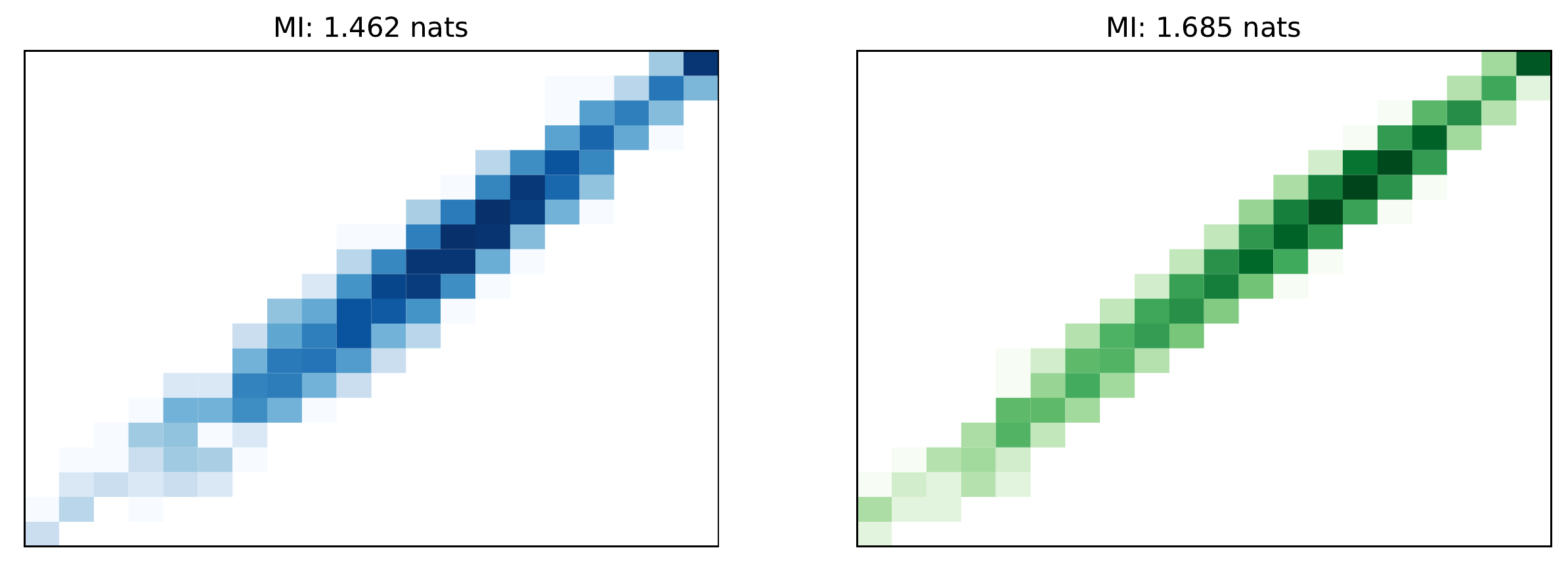}
\label{fig:subfig7}}
\subfloat{
\includegraphics[width=1.0\textwidth]{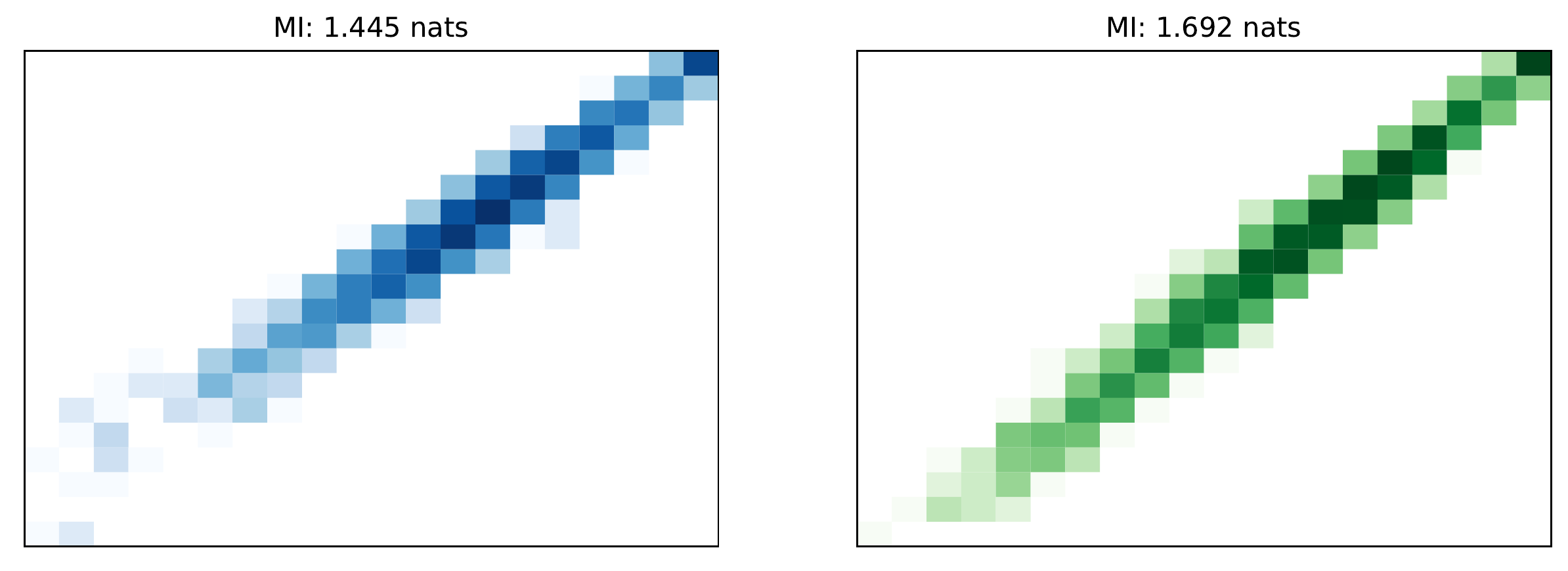}
\label{fig:subfig8}}
\subfloat{
\includegraphics[width=1.0\textwidth]{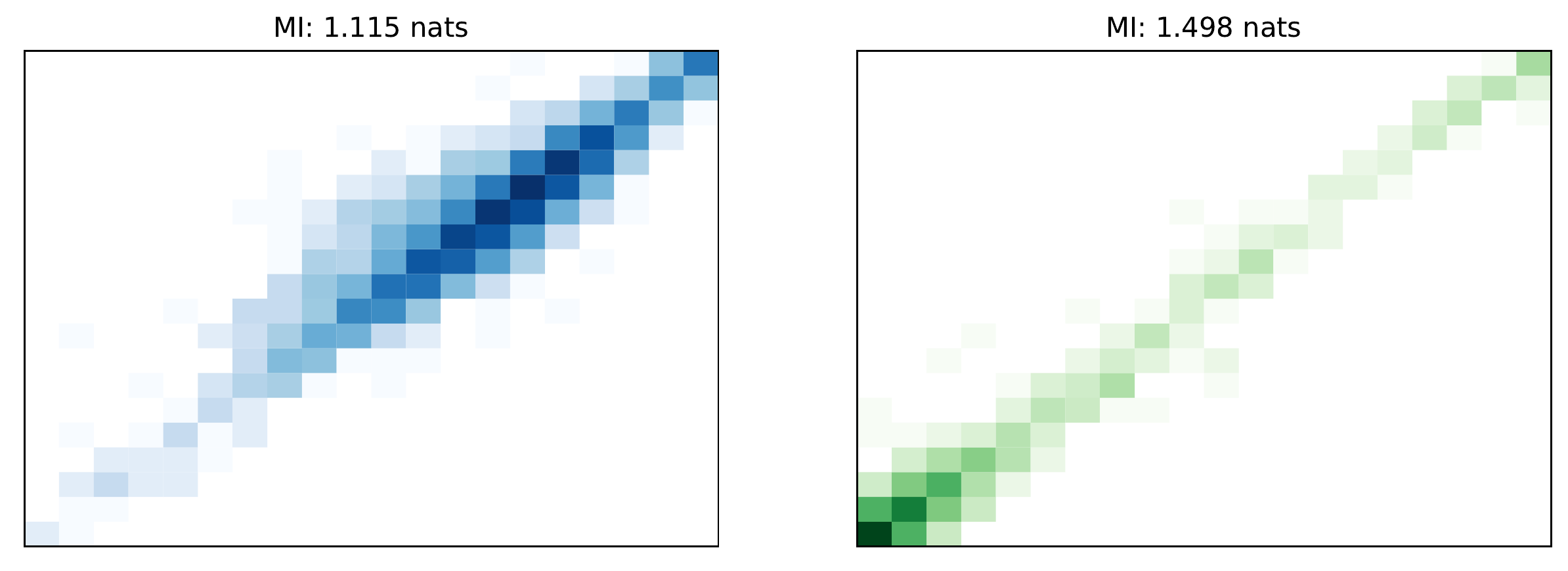}
\label{fig:subfig9}} \\
\subfloat{
\includegraphics[width=1.0\textwidth]{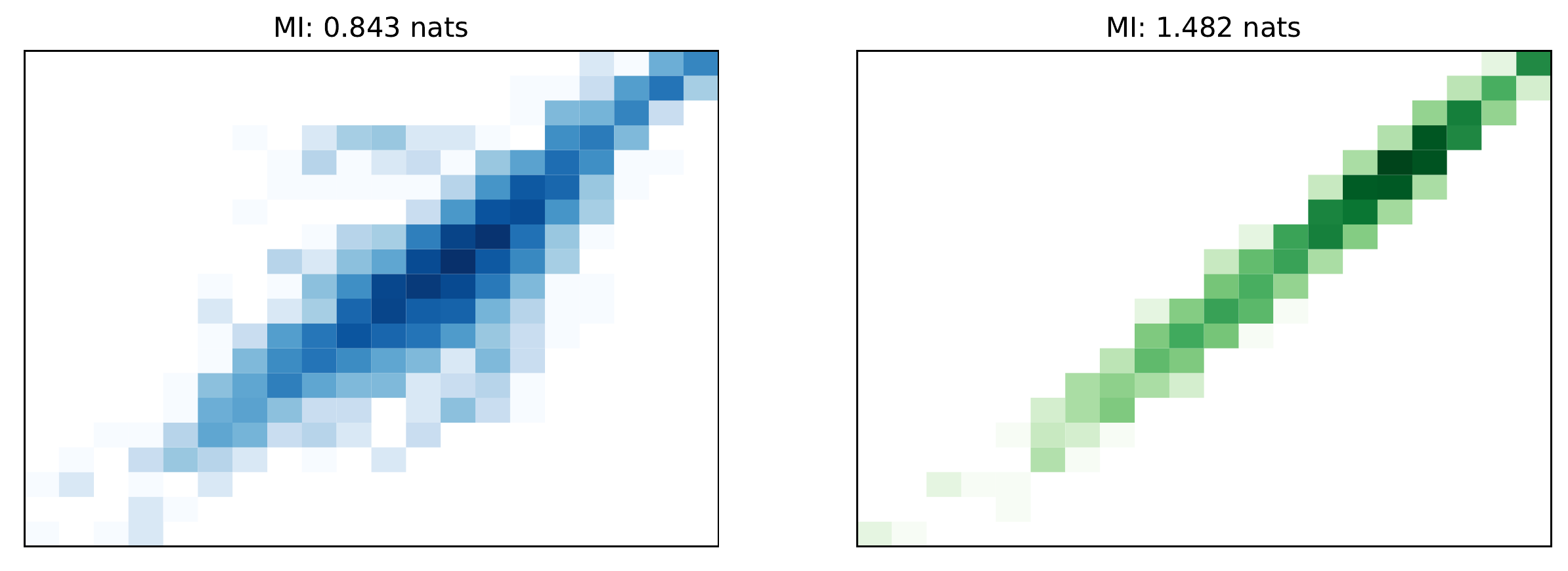}
\label{fig:subfig10}} 
\subfloat{
\includegraphics[width=1.0\textwidth]{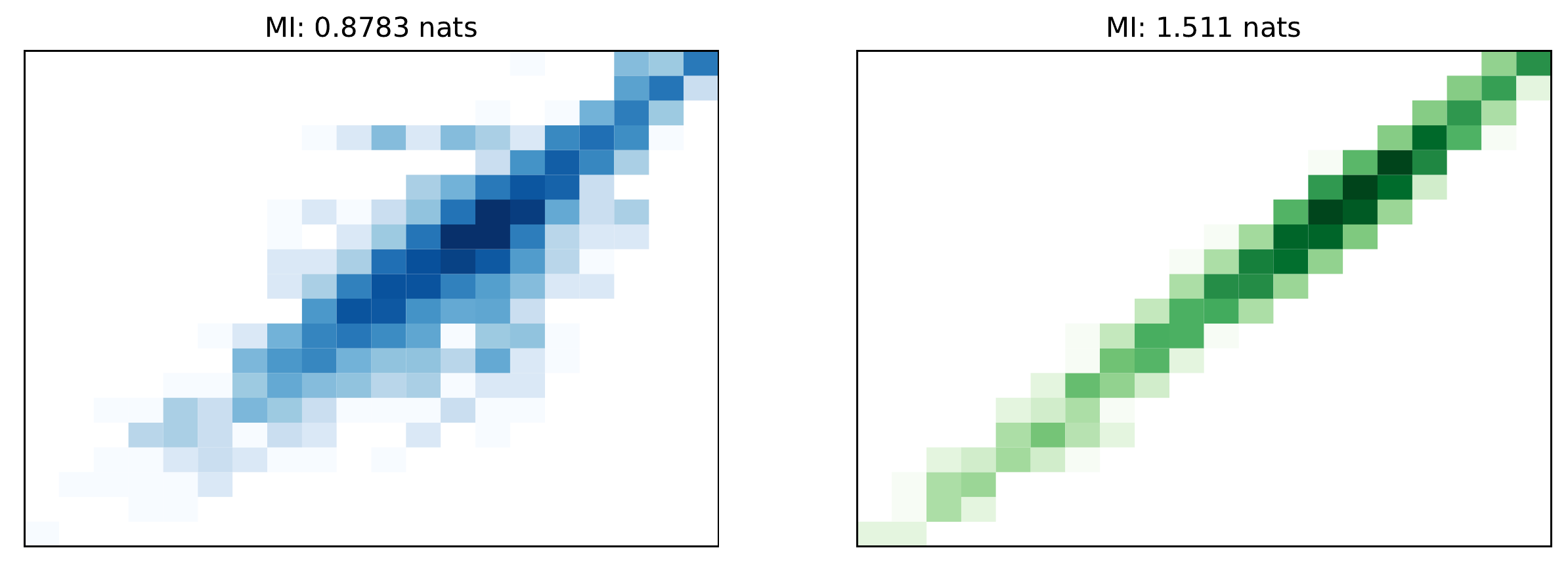}
\label{fig:subfig11}} 
\subfloat{
\includegraphics[width=1.0\textwidth]{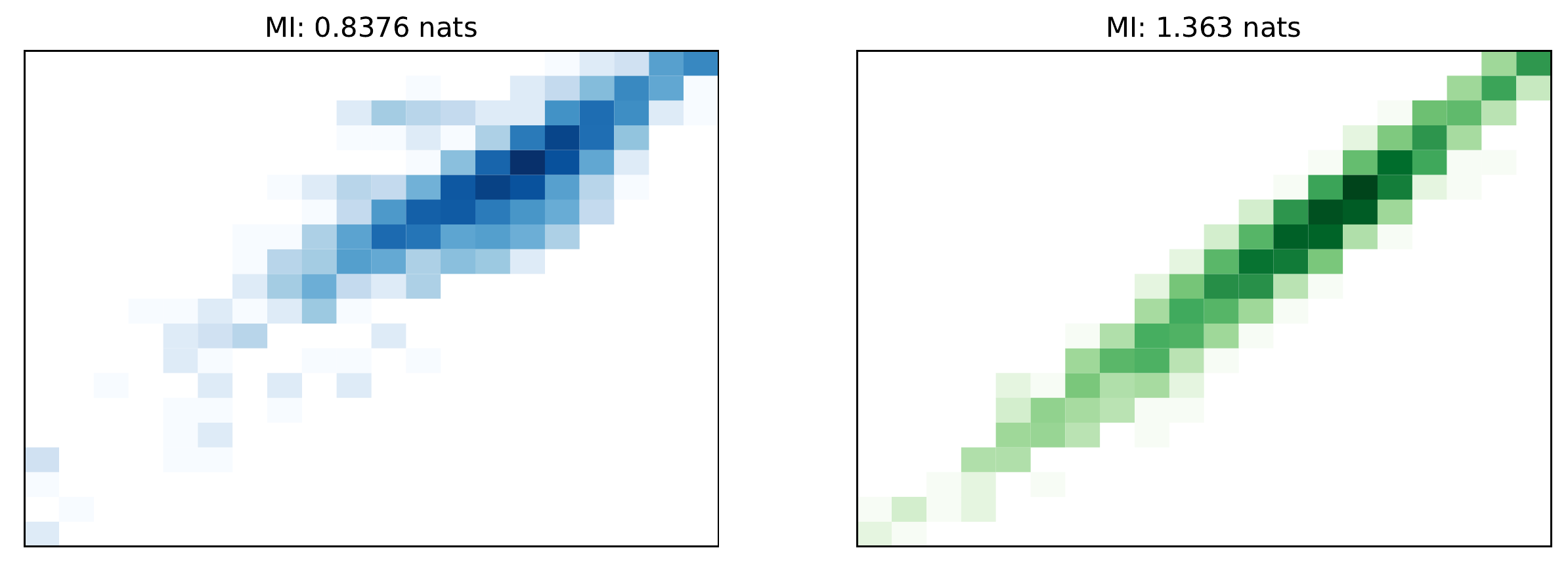}
\label{fig:subfig12}} \\
\end{minipage}
\caption{Joint distribution between two augmentation induced views. Images depict 12 attention slices per methods, obtained by slicing the attention tensor for the input sentence ``the best thing you can do is to know your stuff.'' Increasing depth in layer stack from left to right, top to bottom. (\textcolor{blue}{\fullcirc}) SimCSE, (\textcolor{green}{\fullcirc}): \ours (best viewed in color)}
\label{fig:globfig}
\end{figure*}

\section{Detailed Comparison with SimCSE}\label{sec:non-contrastive}

\begin{figure*}\centering
    \includegraphics[width=0.70\textwidth]{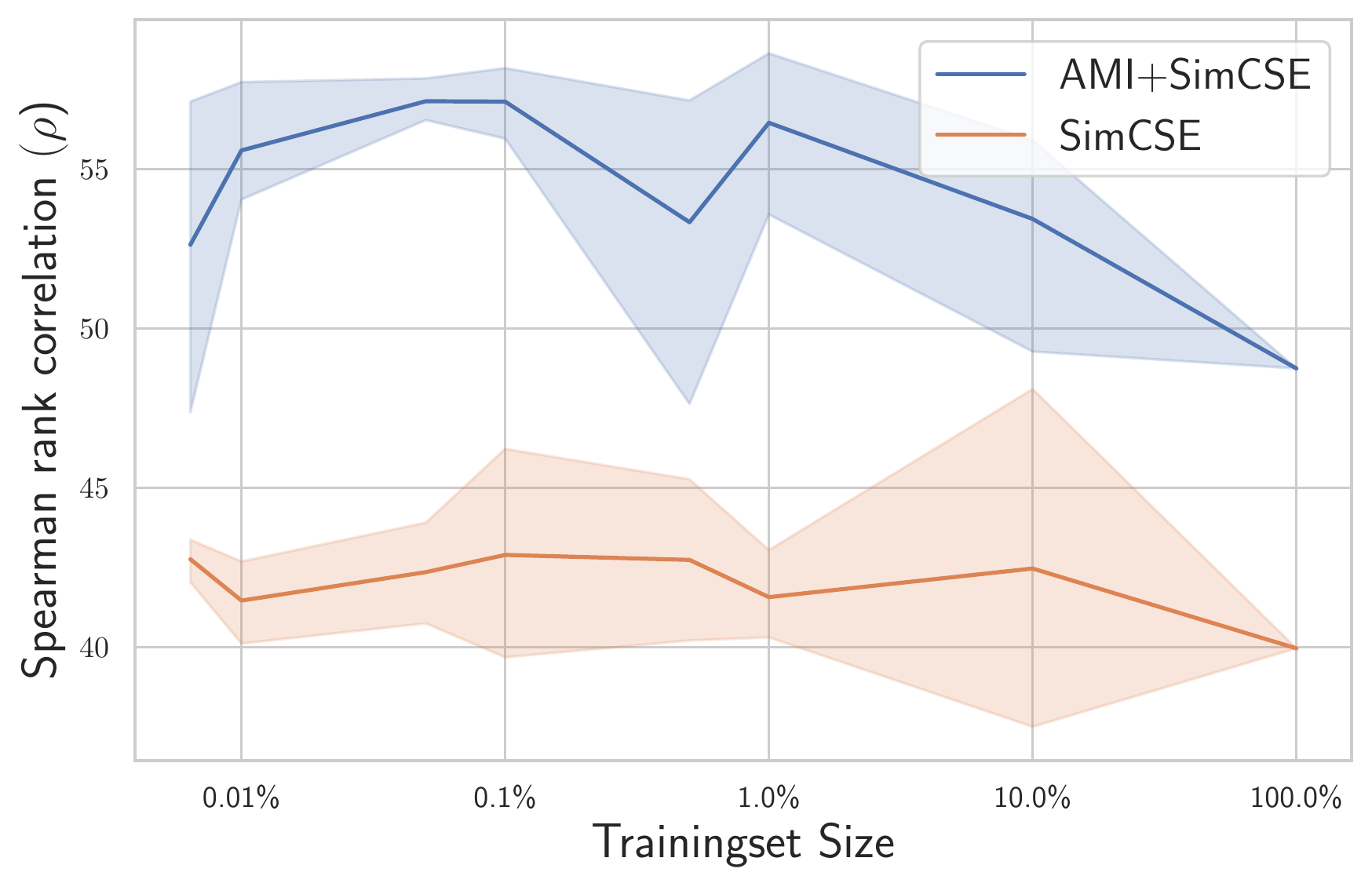}
  
    \caption{Few-shot performance of SimCSE~\cite{gao2021simcse} (\protect\redline) and the proposed approach AMI in combination with SimCSE (\protect\blueline). Performance is shown in Spearman’s correlation average of the STS benchmark at different ratios of dataset sizes used for training. Training in non-contrastive setting with positive-only pairs.} 
    \label{fig:detailed-contrastive-vs-nonconstrative}
\end{figure*}

Our proposed method is built on top of contrastive learning. Thus it intrinsically relies on the existence of the negative pairs.
To complement the performance comparison of contrastive learning in Fig.~\ref{fig:detailed-low-shot-simcse}, we designed an experiment to analyze the extent to which attention regularization alone (AMI) can compensate for the lack of negative pairs. To that end, we conducted training with positive pairs only. See Tab. \ref{tab:sts-few-shot-non-contrastive} and Fig.~\ref{fig:detailed-contrastive-vs-nonconstrative} for results. The integration of mutual attention information boosts the performance by up to $(+15)$ across all training set sizes. 
It suggests the potential application of our proposed attention regularization for non-contrastive learning.

\section{Bivariate Normal Mutual Information}
\label{sec:bivariate_mi}

\noindent\textbf{General Log-Normal Properties: } Similar to the normal distribution, the log-normal distribution $\log \mathcal{N}(w|\mu_w,\sigma^2_w)$ has two parameters $\mu_w$ and $\sigma_w$ capturing mean and variance. It follows that applying the $\log$ transformation on a random variable $w$, we yield random variable $z = \log(w)$, which is normally distributed: $z \sim \mathcal{N}(\mu_z,\sigma^2_z)$.

\noindent\textbf{Mutual Information: }Given a vectors of tuples  $(X_1,X_2 )$ containing i.i.d. points sampled  the joint bivariate normal  distribution of $p(A,B) = \mathcal{N}(\bm{\mu},\Sigma)$ with $\bm{\mu} \in \mathbb{R}^2, \Sigma \in \mathbb{R}^{2\times 2}$.
It can be shown that there exists an exact relationship between mutual information and the correlation coefficient $\rho$~\cite{MI1957} derived from $X_1$ and $X_2$. To that end, we expand the notation:
\begin{equation}
\bm{\mu}=\begin{pmatrix}\mu_\viewA & \mu_\viewB \end{pmatrix}, \quad \Sigma = \begin{pmatrix} \sigma_\viewA^2 & \rho\sigma_\viewA\sigma_\viewB \\ \rho\sigma_\viewA\sigma_\viewB & \sigma^2_\viewB \end{pmatrix}
\end{equation}

The marginal and the joint entropy terms for Gaussian distributed variables can be written as:
\begin{equation}
\label{eq:entropies1}
\begin{aligned}
H(X_i) = \frac{1}{2}\log (2\pi\bm{e}\sigma_i^2) = \\
\frac{1}{2} + \frac{1}{2}\log(2\pi) + \log(\sigma_i),\quad i \in\{1,2\} 
\end{aligned}
\end{equation}
\begin{equation}
\label{eq:entropies2}
\begin{aligned}
H(X_1, X_2) = \frac{1}{2} \log \left[(2\pi\bm{e})^2|\Sigma| \right] = \\ 
1 + \log(2\pi)+\log(\sigma_\viewA\sigma_\viewB)+\frac{1}{2}(1-\rho^2).
\end{aligned}
\end{equation}
Given that Mutual Information can be written in terms of entropy as:

\begin{equation}
     I(X_1,X_2)=H(X_1)+H(X_2)-H(X_1, X_2) 
\label{eq:mutual_information2}
\end{equation}
Then it follows by inserting Eq.~\ref{eq:entropies1},\ref{eq:entropies2} in Eq.~\ref{eq:mutual_information2}:
\begin{equation}
    I(X_1,X_2)=-\frac{1}{2}(1-\rho^2)
\end{equation}

\begin{table*}\centering
\begin{tabular}{lcccc}
   \toprule

        \multicolumn{5}{c}{\it{Semantic Textual Similarity}}\\
         \midrule
Model               & \tf{0.1\%}        & \tf{1\%}          & \tf{10\%}         & \tf{100\%} \\
    \midrule
SimCSE (with negatives)         & $66.69 \pm 1.03$ & $74.08 \pm 0.81$ & $75.01 \pm 0.23$ & $76.15$ \\
$*$ \ours (with negatives)      & ${73.85 \pm 0.49}$ & ${76.21 \pm 0.28}$ & ${76.31 \pm 0.46}$ & ${78.13}$ \\
     \hdashline
SimCSE (w/o negatives)             & $43.02 \pm 4.48$ & $41.30 \pm 1.63$ & $42.56 \pm 6.87$ & $40.18$ \\
$*$ \ours (w/o negatives)        & $57.00 \pm 1.32$ & $56.41 \pm 3.38$ & $53.38 \pm 4.70$  & $54.34$ \\
\bottomrule
\end{tabular}
\caption{Sentence embedding few-shot learning performance on STS tasks measured as Spearman’s correlation. Top: performance in contrastive setup with in-batch negatives. Bottom: performance with positive samples only. The number corresponds to the average performance across all benchmarks.}
\label{tab:sts-few-shot-non-contrastive}
\end{table*}

\end{document}